\documentclass[10pt,journal,compsoc]{IEEEtran}

\usepackage[nocompress]{cite}
\usepackage[pdftex]{graphicx}
\usepackage{amsmath}
\usepackage{algorithmic}
\usepackage{array}
\usepackage{csquotes}
\usepackage{url}
\PassOptionsToPackage{hyphens}{url}\usepackage[hidelinks,breaklinks=true]{hyperref}
\usepackage{booktabs}
\usepackage{caption}
\usepackage{subcaption}
\usepackage{todonotes}
\usepackage{dblfloatfix}
\usepackage{xcolor}

\begin{document}

\title{How Do Neural Networks Estimate Optical Flow? A Neuropsychology-Inspired Study}

\author{David~B.~de~Jong~\href{https://orcid.org/0000-0003-2088-2146}{\includegraphics[scale=0.08]{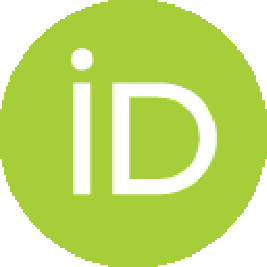}},
        Federico~Paredes-Vall\'es~\href{https://orcid.org/0000-0002-9478-7195}{\includegraphics[scale=0.08]{images/orcid_128x128.eps}},
        and~Guido~C.~H.~E.~de~Croon~\href{https://orcid.org/0000-0001-8265-1496}{\includegraphics[scale=0.08]{images/orcid_128x128.eps}},~\IEEEmembership{Member,~IEEE}
        
    \IEEEcompsocitemizethanks{\IEEEcompsocthanksitem The authors are with the Micro Air Vehicle Laboratory (MAVLab), Department of Control and Simulation, Faculty of Aerospace Engineering, Delft University of Technology, Kluyverweg 1, 2629 HS Delft, The Netherlands. E-mail: \href{mailto:daviddejong@me.com}{daviddejong@me.com}, $\{$\href{mailto:f.paredesvalles@tudelft.nl}{f.paredesvalles}, \href{mailto:g.c.h.e.decroon@tudelft.nl}{g.c.h.e.decroon$\}$@tudelft.nl}\protect}%
}

%\markboth{IEEE Transactions on Pattern Analysis and Machine Intelligence, 2020}%
%{de Jong \MakeLowercase{\textit{et al.}}: How Do Neural Networks Estimate Optical Flow? A Neuropsychology-Inspired Study}

\IEEEtitleabstractindextext{%
\begin{abstract}
End-to-end trained convolutional neural networks have led to a breakthrough in optical flow estimation. The most recent advances focus on improving the optical flow estimation by improving the architecture and setting a new benchmark on the publicly available MPI-Sintel dataset. Instead, in this article, we investigate how deep neural networks estimate optical flow. A better understanding of how these networks function is important for (i) assessing their generalization capabilities to unseen inputs, and (ii) suggesting changes to improve their performance. For our investigation, we focus on FlowNetS, as it is the prototype of an encoder-decoder neural network for optical flow estimation. Furthermore, we use a filter identification method that has played a major role in uncovering the motion filters present in animal brains in neuropsychological research. The method shows that the filters in the deepest layer of FlowNetS are sensitive to a variety of motion patterns. Not only do we find translation filters, as demonstrated in animal brains, but thanks to the easier measurements in artificial neural networks, we even unveil dilation, rotation, and occlusion filters. Furthermore, we find similarities in the refinement part of the network and the perceptual filling-in process which occurs in the mammal primary visual cortex.
\end{abstract}

}

\maketitle
\IEEEdisplaynontitleabstractindextext
\IEEEpeerreviewmaketitle

\IEEEraisesectionheading{\section{Introduction}\label{sec:introduction}}

\IEEEPARstart{O}{ptical flow} is a visual cue defined as the projection of the apparent motion of objects in a scene onto the image plane of a biological vision system or a visual sensor \cite{Gibson1950}. This cue is important for the behavior of animals of varying size \cite{Feng2003}, ranging from small flying insects \cite{Borst2010} to humans \cite{Borst2015}, as it allows these animals to estimate their ego-motion and to have a better understanding of the visual scene. Optical flow is also important in computer vision and robotics applications for tasks such as object tracking \cite{Kajo2017} and autonomous navigation \cite{de2016monocular}.

Many algorithms have been introduced to determine optical flow \cite{beauchemin1995computation}, including correlation-based matching methods \cite{Anandan1989,Singh1991}, frequency-based methods \cite{Heeger1988,Fleet1992}, and differential methods \cite{Horn1981,Lucas1981}. Correlation-based matching methods try to maximize the similarity between different intensity regions across multiple frames. Finding the best match then corresponds to finding the shift which maximizes the similarity score. Frequency-based methods exploit either the amplitude or phase component of the complex valued response of a Gabor quadrature filter pair \cite{Gabor1945} convolved with an image sequence. Lastly, differential methods compute optical flow based on a Taylor expansion of the image signal, subject to the brightness constancy assumption.

All these methods assume that the brightness of a moving pixel remains constant over time and, when applied locally, are subject to the \textit{aperture problem} \cite{Ullman1979}. Only motion components normal to the orientation of an edge in the image can be resolved.

A global smoothness constraint has been added for differential methods, which assumes that neighboring pixels undergo a similar motion \cite{Horn1981}. This has led to \emph{variational} methods that minimize a global energy function consisting of a data and a smoothness term. These methods have played a dominant role for many years due to their high performance. However, a main drawback is that the iterative minimization of the energy function leads to long computation times. Moreover, the brightness constancy assumption is a coarse approximation to reality and thus limits performance \cite{Zimmer2011}. Research has focused on extra energy terms to deal with deviations from the brightness constancy assumption and improve the robustness of global smoothness constraints, leading to slow but steady progress.

As in many other computer vision areas, currently, the best-performing algorithms are trained deep neural networks. Initially, training such networks was challenging due to the lack of ground-truth optical flow data and the excessive human effort required for manual optical flow labeling. \textit{Dosovitskiy et al.} \cite{Dosovitskiy2015} were the first to successfully train deep neural networks to estimate optical flow by using a synthetically generated dataset with optical flow ground truth. Their networks, FlowNetS and FlowNetC, initially performed slightly worse than the state-of-the-art variational methods \cite{Tu2019}. However, trained deep neural networks became the new state-of-the-art method for optical flow estimation by subsequent researchers who focused on improving the architecture and training data \cite{Ilg2017,Sun2018,Hui2019}. 

Until now, the functioning of these networks is poorly understood. In this article we investigate \textit{how} deep neural networks perform optical flow estimation. Besides satisfying curiosity, there are two main reasons why this is important. First, understanding the method's functioning brings insights into its limits and robustness, for example concerning generalization to test distributions. Second, it may lead to valuable recommendations for improving the performance, for instance, by changing properties of the architecture or training data. 

In our analysis of deep optical flow networks, we make use of a method that has helped unveiling the workings of motion-sensitive brain areas in neuropsychology \cite{Jones1987a}. Specifically, we measure the response of neurons in FlowNetS \cite{Dosovitskiy2015} to stimuli with varying spatiotemporal frequencies and construct a spectral response profile. The input stimuli used are translating plane waves, as this input type proved to be more selective in the frequency domain than moving bars \cite{Albrecht1980}. Based on the earlier findings of Gabor filters \cite{Gabor1945} in biological vision systems \cite{Jones1987,DeAngelis1995} and other learning-based methods \cite{VanHateren1998, Olshausen2003}, we expect to find these filters in FlowNetS as well. Therefore, we fit a Gabor function to the spectral response profile of neurons in the network and study the residual error patterns. We find that the Gabor translational motion filter model is suitable for the majority of the neurons. Additionally, we find neurons sensitive to motion patterns such as dilation, rotation, and occlusion. Interestingly, neurons sensitive to these motion patterns have not been mentioned in neuropsychology. Furthermore, our analysis strongly suggests that the resolution in the temporal frequency domain can be significantly improved if more than two frames would be used as input to the neural network. Lastly, we find that the optical flow refinement process in the decoder part of the network behaves similarly in function to flow refinement in biological vision systems.

\begin{figure}[!t]
	\centering
	\includegraphics[width=0.49\textwidth]{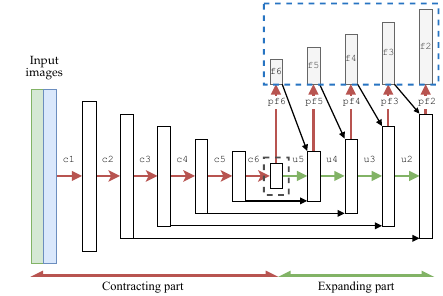}
	\caption{Schematic of the FlowNetS architecture \cite{Dosovitskiy2015}. The contracting part compresses spatial information through the use of strided convolutions (\texttt{c}), while the expanding part uses upconvolutions (\texttt{u}) for refinement. The predict-flow (\texttt{pf}) layers transform feature map activations into dense flow estimates (\texttt{f}). 
		The feature map corresponding to the output of the \texttt{c6} layer (gray dashed box) is studied in Sections \ref{sec:gabfit} \& \ref{sec:dilrot}, while the flow refinement process (blue dashed box) is discussed in Section \ref{sec:ap}. 
		}
	\label{fig:archflownet}
\end{figure}

The remainder of the article is structured as follows. In Section \ref{sec:rel}, related work in neuropsychology and deep-learning is discussed. In Section \ref{sec:bg}, an explanation is given of the architecture of FlowNetS (see Fig. \ref{fig:archflownet}). In Section \ref{sec:gabfit}, the network's neural responses to translating wave patterns are studied, and compared to translational Gabor filters. Subsequently, Section \ref{sec:dilrot} discusses the response of neurons to dilating and rotating waves. In Section \ref{sec:ap}, it is studied how FlowNetS resolves the aperture problem. Finally, the results of this work are discussed in Section \ref{sec:dis}, and conclusions are drawn in Section \ref{sec:conc}.

\section{Related Work} \label{sec:rel}

\subsection{Dense optical flow estimation with CNNs}
Ever since the pioneering work of \textit{Horn et al.} \cite{Horn1981}, variational optical flow methods \cite{Brox2004} have played a dominant role in optical flow estimation due to their high performance. Most modern variational optical flow estimation pipelines consist of four stages: matching, filtering, interpolation, and variational refinement. Various improvements have been proposed over time to deal with issues such as long-range matching \cite{Brox2011} and occlusion \cite{Revaud2015}. Furthermore, improvements such as dense correspondence matching based on convolution response maps of the reference image with the target image \cite{Weinzaepfel2013}, and supervised data-driven interpolation of a sparse optical flow map \cite{Zweig2017} were also proposed. These last two improvements introduced elements of deep learning into the variational optical flow estimation pipeline. 

\textit{Dosovitskiy et al.} \cite{Dosovitskiy2015}, however, were the first to introduce a supervised end-to-end trained Convolutional Neural Network (CNN). CNNs have three major advantages when it comes to estimating optical flow. First, CNNs outperform variational optical flow estimation methods in terms of accuracy \cite{Ilg2017,Sun2018,Hui2019}. Second, the runtime of CNN-based optical flow algorithms, when executed on the appropriate hardware, is significantly lower than variational methods \cite{Ilg2017}. Third, CNN-based methods can learn from data and can exploit statistical patterns not realized by a human designer. This is an advantage over variational methods which require explicit, and sometimes inaccurate, assumptions on the input. However, CNNs also have three disadvantages. First, the results depend on the quality and size of the training data. Second, CNN-based methods face the risk of overfitting, which is relevant for optical flow estimation because it is difficult to obtain ground truth \cite{Tu2019}. Third, there is no guarantee that the trained models will generalize to scenarios not contained in the training dataset. Due to the \emph{``black-box''} nature of the solution, it is difficult to get insight into its workings and limitations.

In \cite{Dosovitskiy2015}, \textit{Dosovitskiy et al.} introduced two networks based on the U-net architecture \cite{Ronneberger2015}: FlowNetS and FlowNetC. While FlowNetS is an encoder-decoder network consisting of simple convolutions, FlowNetC creates two separate processing streams and combines them in a \textit{correlation-layer}. This layer performs a multiplicative patch comparison between feature maps. Due to the explicit use of a correlation-layer, it is more straightforward to understand the workings of FlowNetC. However, not much is known about the workings of FlowNetS. Inspired by this architecture, \textit{Ranjan et al.} \cite{Ranjan2017a} introduced SpyNet, a spatial image pyramid with simple convolutional layers at each pyramid level and a warping operation between pyramid levels. SpyNet's coarse-to-fine approach brings a higher computational and memory efficiency at the cost of a more limited set of perceivable motion types. \textit{Ranjan et al.} also visualized the weights of the first layer of their network and observe that these filters resemble Gabor filters \cite{Gabor1945}, which provided a glimpse into the working principle of this architecture. Finally, \textit{Teney et al.} \cite{Teney2016} built a shallow CNN-architecture by integrating domain knowledge, such as invariance to brightness and  in-plane rotations. On small motion their architecture performs well, but performance declines on large motion near occlusions. They conclude good occlusion performance requires reasoning over a larger spatiotemporal extent, which their shallow architecture is not able to do.

The generalization performance of CNN-based methods can be evaluated for specific instances by determining the epistemic uncertainty \cite{kendall2017uncertainties}. Indeed, \textit{Ilg et al.} \cite{Ilg2018b} used a modified FlowNetC that produces multiple hypotheses per forward pass, which are then merged to a single distributional flow output. They showed that their network produces highly uncertain flow estimates when optical flow estimation is difficult (shadows, translucency, etc.). Lastly, \textit{Ranjan et al.} \cite{Ranjan2019} highlighted another downside of deep neural networks, which is the ability of adversarial examples to fool neural networks and produce erroneous results. They showed that especially networks using an encoder-decoder architecture are affected, while networks using a spatial pyramid framework are less vulnerable. None of the works above, however, explain how their architecture estimates optical flow.

\subsection{Receptive field mapping}
There are two main threads of research to understand what neural networks have learned: attribution and feature visualization. Attribution methods \cite{Springenberg2014,Zeiler2014} are used to \textit{attribute} filter outputs, like optical flow, to parts of the input by visualizing the gradient. However, it is hard to see where an optical flow estimate comes from. Feature visualization is concerned with understanding what neurons, filters, or layers in a neural network are sensitive to by optimizing the input \cite{Erhan2009}. The result is usually an image with noisy and visually difficult to interpret high-frequency patterns \cite{Olah2017}. Three methods of regularization can be applied to cope with this phenomenon. First, frequency penalization discourages the forming of these patterns. The downside is that this approach also discourages the forming of legitimate high-frequency patterns which are of interest for optical flow estimation. Second, small transformations like scaling, rotation, or translation can be applied in between optimization steps \cite{Mordvintsev2015}. This approach is also not viable because transformation affects the ground truth of optical flow. Third, priors can be used which can keep the optimized input interpretable. Such approaches typically involve learning a generative model \cite{Nguyen2016a} or enforcing priors based on statistics from the training data \cite{Wei2015}. This approach is often very complex and it may be unclear what can be attributed to the prior and what can be attributed to what the network has learned.

Due to these reasons, we look at the field of neuropsychology and specifically study what methods researchers have used to determine what stimuli activate neurons in mammalian vision systems and what functions best describe the neural responses. It was shown that Gabor functions \cite{Gabor1945} best modeled the spatial response of simple cells in the mammal visual cortex \cite{Jones1987}. It can be shown that Gabor filters are optimal for simultaneously localizing a signal in the spatial and frequency domain \cite{Bracewell1986}, making them ideal for motion estimation. Later, \textit{DeAngelis et al.} \cite{Deangelis1993} examined the spatiotemporal response of cells and their space-time separability. In functional form, space-time separable Gabor filters are frequency-tuned with a stationary Gaussian envelope and space-time inseparable Gabor filters are velocity-tuned with a moving Gaussian envelope  \cite{Petkov2007}. In this work we only consider fitting frequency-tuned Gabor filters, due to their simplicity and the low number of input frames used by the FlowNet architectures.

Two approaches to receptive field mapping in neuropsychology can be discerned: the reverse-correlation approach and the spectral response profile approach. The former presents a rapid random sequence of flashing bars at various imaging locations to the mammal. The spike train emitted by the neuron in the subject is correlated to the sequence in which the stimuli were presented. This approach allows for a rapid measurement of the receptive field profile in the spatiotemporal domain \cite{DeAngelis1995}. Instead, the spectral response profile approach presents translating plane waves to the mammal at varying orientations and spatiotemporal frequencies \cite{Palmer1981,DeAngelis1993a}. \textit{Jones et al.} used both the reverse-correlation approach to construct a spatial receptive field profile \cite{Jones1987b} and measured the response to plane waves to construct a spectral response profile \cite{Jones1987a}. Subsequently, the spatial and spectral responses were compared to the Gabor filter model in the spatial and frequency domain, and the filter parameters obtained from both methods proved to be highly correlated \cite{Jones1987}. A similar correspondence in outcome between the methods was found by \textit{Deangelis et al.} \cite{Deangelis1993,DeAngelis1993a} in the visual cortex of cats.

In this work we extend the approach of \textit{Jones et al.} \cite{Jones1987} to the spatiotemporal domain and measure spectral responses of the network to translating plane waves, to which frequency-tuned spatiotemporal Gabor filters are fitted. A benefit of measuring the spatiotemporal spectral responses for optical flow is that translation is more easily described in the frequency domain \cite{Petkov2007}. 

\subsection{Aperture problem}
Optical flow estimations methods are only able to resolve motion components normal to the orientation of  an edge in the intensity pattern. This is known as the aperture problem \cite{Ullman1979}. In CNNs the size of the aperture of a neuron is referred to as the receptive field, which is defined as the region in the input which affects the activation of the neuron. In this work we show that the receptive field size is related to the aperture problem by training different versions of FlowNetS with varying receptive field sizes.

In neuropsychology, \textit{Komatsu} \cite{Komatsu2006} has shown the existence of a perceptual filling-in mechanism in the mammalian visual cortex for cues such as color, brightness, texture, or motion. While the precise neural workings are still under discussion, edge structure \cite{VonDerHeydt2003} and the interaction between neighboring neurons play an important role in this process \cite{Poort2012}. In neural networks attempts have been made to implement such a mechanism as well. To allow for the interaction between neurons, a recurrent model can be used \cite{Liang2015}. \textit{Zweig et al.} \cite{Zweig2017}, however, used an unfolded feed-forward version of a recurrent network and a multi-layer loss to allow for interaction between neurons. Their CNN-based motion interpolation architecture takes a sparse flow map and edge structure as input. They showed their motion interpolation method refines motion estimates similarly to the human visual cortex by demonstrating the filling-in effect of the network on a Kanizsa illusion. FlowNetS also features a multi-layer loss, and, in Section \ref{sec:ap}, the ability of the expanding part of FlowNetS to interpolate and refine flow maps is highlighted.

\section{Model details}\label{sec:bg}

Fig. \ref{fig:archflownet} shows a schematic representation of the FlowNetS architecture, which takes two consecutive images as input. Multiple versions of FlowNetS exist. \textit{Dosovitskiy et al.} \cite{Dosovitskiy2015} mention the use of the ReLU activation function in their work. The release of their pre-trained models, however, uses a leakyReLU activation function\footnote{\url{https://lmb.informatik.uni-freiburg.de/resources/binaries/flownet/flownet-release-1.0.tar.gz}}. In order to facilitate interpretability of the motion filter analysis, we choose to use the ReLU version. With the same aim, we introduce two small adjustments. First, the bias terms are removed in the predict-flow \texttt{pf} layers because the flow is assumed to be zero-centered. Second, the kernel size in the \texttt{pf} layers is reduced from $3 \times 3$ to $1 \times 1$ to allow clearer location identification. The full details of our version of FlowNetS can be found in Appendix A. 

Regarding training, as in \cite{Dosovitskiy2015}, we use the same data augmentation \textit{on both} frames, but we do not use incremental flow and color augmentation \textit{between} frames, since the authors do not specify the parameters of these mechanisms. Furthermore, the network is trained for fewer iterations (300K iterations versus 600K iterations) due to limited availability of computational resources. Evaluation on the MPI-Sintel \cite{Butler2012} and FlyingChairs \cite{Dosovitskiy2015} datasets shows comparable performance between the slightly modified FlowNetS and the original version, as can be seen in Appendix A.

The synthetic dataset FlyingChairs \cite{Dosovitskiy2015}, which was used to train the original and our slightly modified FlowNetS, consists of approximately 22k image pairs. The image pairs are composed of a varying numbers of chairs and background images from natural scenes. Between image pairs, a composition of translation, rotation, and scaling motion is applied. As stated in the supplementary material of \cite{Dosovitskiy2015}, the size of the chairs\footnote{Note that, in \cite{Dosovitskiy2015}, the authors do not specify how the size of a chair is determined, so there is a certain ambiguity around this parameter.} is sampled from a Gaussian with a mean and standard deviation of 200 pixels, clamped between 50 and 640 pixels. Note that the synthetic scenes also contain occlusion. Further details about the composition of affine motion can be found in \cite{Dosovitskiy2015}.

\section{Gabor spectral response profile fitting for translation}\label{sec:gabfit}
We investigate to what motion patterns the neurons in FlowNetS are sensitive. In neuropsychology, the responses of simple cells turned out to be captured very well by Gabor filters \cite{Jones1987,Jones1987a,DeAngelis1993a,touryan2004nonlinear}. 
That simple cells act like Gabor filters makes sense, since Gabor filters are known to be optimal in the sense that they achieve maximal resolution in both the spatiotemporal and the associated frequency domains. As a consequence, they require a minimal number of filters to represent spatiotemporal information \cite{Jones1987,DeAngelis1993a}. 

Although artificial neural networks are very different in many aspects from biological ones, they were inspired by them and inherit similar traits. In particular, they seem suitable to represent spatiotemporal filters and may be subject to a similar pressure as biological networks to succinctly represent spatiotemporal patterns when having to estimate optical flow. This was our motivation to first investigate whether FlowNetS' neural responses resemble those of Gabor filters. In our investigation, we mainly focus on the deepest encoding layer in the network, the \texttt{c6} layer. As shown in Fig. \ref{fig:archflownet}, the activations of the feature maps of these layers are directly, linearly transformed (via \texttt{pf6}) into an initial coarse-scale horizontal and vertical flow estimate (i.e. \texttt{f6}), which is later used as the basis for refinement. Hence, the coarsest, most direct representation of optical flow is encoded in this layer. Although we focus our analysis on \texttt{c6}, the earlier layers play an important role as well. They do this not only by the determination of the activations in layer \texttt{c6} but also (in the case of \texttt{c2} - \texttt{c5}) by contributing to the refinement of optical flow via skip connections. 

In this section, first the theory behind Gabor filters and the spectral response fitting method is discussed, followed by the results obtained. Thereafter, we discuss the resolution in the temporal frequency domain of the fitted Gabor filters.

\subsection{Methodology}

As in \cite{Gabor1945,Heeger1988,Petkov2007}, the spatiotemporal frequency-tuned Gabor filter $g$ in Cartesian coordinates centered at the origin can be written as the product of a Gaussian $w$ and a translating plane wave $s$:
\begin{equation}
g(x, y,t)=s(x, y,t) w(x, y,t)
\end{equation}

The (non-normalized) Gaussian $w$ is defined by:
\begin{equation}
w(x,y,t) = \exp \left( -  \frac{1}{2}\left( \frac{x_R^2}{\sigma_x^2} + \frac{y_R^2}{\sigma_y^2}  + \frac{t^2}{\sigma_t^2}\right) \right)
\end{equation}
\noindent where $\sigma_x$, $\sigma_y$, and $\sigma_t$ control the spread of the spatiotemporal Gaussian window. To decrease the number of parameters in the fitting process, it is assumed that the center of the Gaussian coincides with the center pixel of the receptive field. Furthermore, the subscript $R$ denotes a rotation operation which allows the Gaussian to be aligned along orientation $\theta_0$, and is defined as:
\begin{equation} \label{x_r}
\begin{array}{l}{x_R=x \cos (\theta_0)+y \sin (\theta_0)} \\ {y_R=-x \sin (\theta_0)+y \cos (\theta_0)}\end{array}
\end{equation}
\noindent where a positive value of $\theta_0$ corresponds to a clockwise rotation with respect to the positive $x$-axis. The subscript $0$ indicates the parameter value corresponding to the peak response of the Gabor filter. This orientation, which corresponds to the preferred direction of motion of the filter, is related to the spatial frequencies via $\theta_0=\tan ^{-1} \left( f_{y_{0}}/ f_{x_{0}}\right)$.

\begin{figure}[!t]
	\centering
	\includegraphics[width=0.375\textwidth]{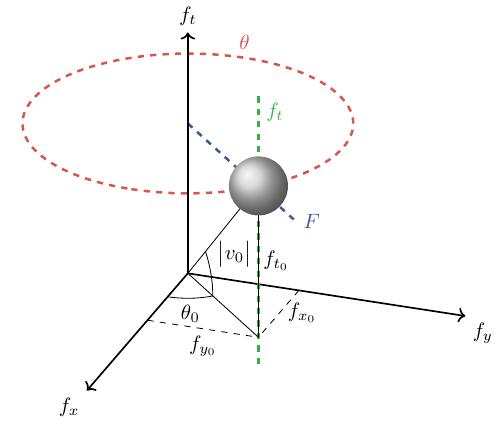}
	\caption{Illustration of the half-magnitude profile in the 3D frequency domain of a spatiotemporal Gabor filter. The three ranges along which the responses of the Gabor half-magnitude profile are evaluated for the spectral response profile fitting process are shown in color.}
	\label{fig:gaborspace}
\end{figure}

A translating plane wave $s$ in the Cartesian coordinate system can be written as:
\begin{equation}
s(x, y, t)=\cos \left(2 \pi  \left( F_{0} x_R -f_{t_{0}}t \right)+\varphi_{0}\right)
\label{eq:trans}
\end{equation}
\noindent where the spatial frequency magnitude $F_0$ is related to the spatial frequencies via $F_0=(f_{x_{0}}^{2}+f_{y_{0}}^{2})^{1/2}$, $f_{t0}$ indicates the temporal frequency, and $\varphi_0$ denotes the phase of the filter. The dependence of $s$ on $y$ is due to $x_R$, which is a function of $x$ and $y$ (see Eq. \ref{x_r}). A Gabor filter is said to be even when $\varphi_0=0$ and odd when $\varphi_0=\pm\pi$. Further, note that the preferred velocity of the filter $v_0$ is related to $F_0$ and the temporal frequency $f_{t_{0}}$ via $v_0=f_{t_{0}}/F_0$, as in \cite{Heeger1988}. A higher spatial frequency $F_0$ allows tracking of motion of thinner image structures. When a signal is sampled in time or space, frequency components which are larger than or equal to $0.5$ cycles per frame (i.e., the Nyquist frequency) become undersampled and aliasing occurs. Thus, if we limit ourselves to signals which do not suffer from aliasing, the maximum velocity a signal can have is limited by its $F_0$. Fig. \ref{fig:gaborspace} shows the 3D frequency space with the half-magnitude profile of a Gabor filter.  

Because we will fit the response of phase-sensitive Gabor filters, we highlight three phase-dependent convolution phenomena. Note that a valid convolution\footnote{We use ``convolution'' to refer to the correlation of a filter over an image to remain consistent with the CNN terminology.} of two tensors with equal size corresponds to their dot product. First, because a sine is an odd signal, the dot product of two sines at opposite frequencies is negative. Second, the dot product of a cosine at opposite frequencies will be positive due to the even nature of the function. Third, sine and cosine are decorrelated and thus the dot product will be zero between these two signals.

\noindent \textbf{Gabor spectral response profile fitting} \\
\noindent In the Gabor spectral response fitting process, translating grayscale plane waves $s$ are used as input to the network, and we try to minimize the difference in response between filters in the \texttt{c6} layer of our FlowNetS and spatiotemporal Gabor filters $g$. To better approximate the response of \texttt{c6} filters, we enhance the Gabor filter output with a gain term $K$, a bias term $b$, and pass the response through a ReLU non-linearity. Then, the response $r$ to a convolution with a translating plane wave $s$ and a Gabor filter $g$ is given by:
\begin{equation}
r = \textnormal{ReLU}\left(K(s(x,y,t) \ast g(x,y,t)) + b\right)
\label{eq:fitgab}
\end{equation}
\noindent where $r$ is a function of nine parameters (i.e., $F_0$, $\theta_0$, $f_{t_{0}}$, $\varphi_0$, $\sigma_x, \sigma_y, \sigma_t$, $K$, $b$), which are estimated in a two-step process.

First, a gridsearch is performed to determine the location in the spatiotemporal frequency domain with the highest response per filter in the \texttt{c6} layer. We denote the response of the filters in the network by $\hat{r}$, and their peak response value by $\hat{r}_0$. Because the fitted Gabor filters are phase sensitive, this amounts to estimating four parameters (i.e., $F_0$, $\theta_0$, $f_{t_{0}}$, $\varphi_0$). Therefore, a four-dimensional grid of translating plane waves (i.e., the input to the network) is constructed using all combinations of these parameters within a given range and step size (see Appendix B). The range for the value of half spatial wavelength $\lambda/2=1/2F$ is chosen so that it captures the sizes of the chairs present in the training dataset (as explained in Section \ref{sec:bg}).

Second, once the peak response of the \texttt{c6} filters is found, we estimate the spatiotemporal spread of the Gaussian (determined by $\sigma_x, \sigma_y, \sigma_t$), the gain $K$, and the bias $b$. This is done by minimizing the difference in response between the fitted Gabor filters $r$ (see Eq. \ref{eq:fitgab}) and the corresponding \texttt{c6} filters $\hat{r}$ along three separate ranges in the spatiotemporal frequency space ($F$, $\theta$, and $f_t$). These ranges are illustrated in Fig. \ref{fig:gaborspace}, and further described in Appendix B. We define the cost function $\mathcal{L}$ in response to a convolution with a translating plane wave $s$ as:

\begin{equation}
\begin{split}
\mathcal{L} &= \sum_{i}^{} (r_i-\hat{r}_i)_{F}^{2} + \sum_{j}^{} (r_j-\hat{r}_j)_{\theta}^{2} + \sum_{k}^{} (r_k-\hat{r}_k)_{f_t}^{2}\\&= \mathcal{L}_F + \mathcal{L}_\theta + \mathcal{L}_{f_{t}}
\label{eq:min}
\end{split}
\end{equation}
\noindent where $\mathcal{L}_F$, $\mathcal{L}_\theta$, and $\mathcal{L}_{f_{t}}$ denote the sum of squared errors over the respective ranges. We constrain the bounds of the Gabor filter parameters to obtain reasonable values, which leads to a non-linear bounded convex optimization problem which is solved using the robust trust-region-reflective algorithm \cite{Iciam2000}. In order to compare the obtained cost values between \texttt{c6} filters, we construct a normalized cost value $\mathcal{L}_{norm}$ by dividing the cost by the squared peak response of the filter: $\mathcal{L}_{norm} = \mathcal{L}/\hat{r}_{0}^2$.

\begin{figure*}[!t]
	\begin{subfigure}[t]{0.5\textwidth}
		\centering
		\includegraphics[width=0.85\textwidth]{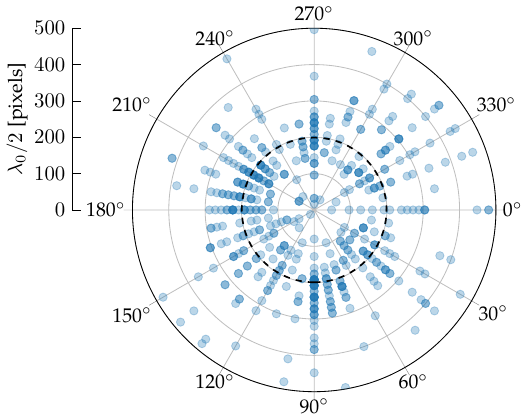}
	\end{subfigure}%
	~ 
	\begin{subfigure}[t]{0.5\textwidth}
		\centering
		\includegraphics[width=0.7\textwidth]{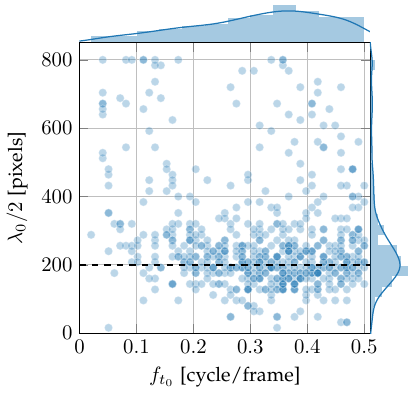}
	\end{subfigure}
	\caption{Location of peak response $\hat{r}_0$ per \texttt{c6} filter in the spatiotemporal frequency domain in response to translating plane waves. \emph{Left:} Half spatial wavelength $\lambda_0/2$ and orientation $\theta_0$ corresponding to peak response $\hat{r}_0$ per filter. \emph{Right:} Half spatial wavelength $\lambda_0/2$ and temporal frequency $f_{t_{0}}$ corresponding to peak response $\hat{r}_0$ per filter. In both plots, the black dashed lines indicate the peak of the distribution in the half spatial wavelength dimension, which is around 200 pixels. 
	}
	\label{fig:trans}
\end{figure*}

\begin{figure*}[!t]
	\begin{subfigure}[t]{0.2\textwidth}
		\centering
		\includegraphics[width=0.9\textwidth]{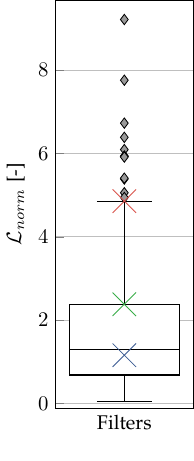}
	\end{subfigure}%
	\centering
	\begin{subfigure}[t]{0.8\textwidth}
		\centering
		\includegraphics[width=1.\linewidth]{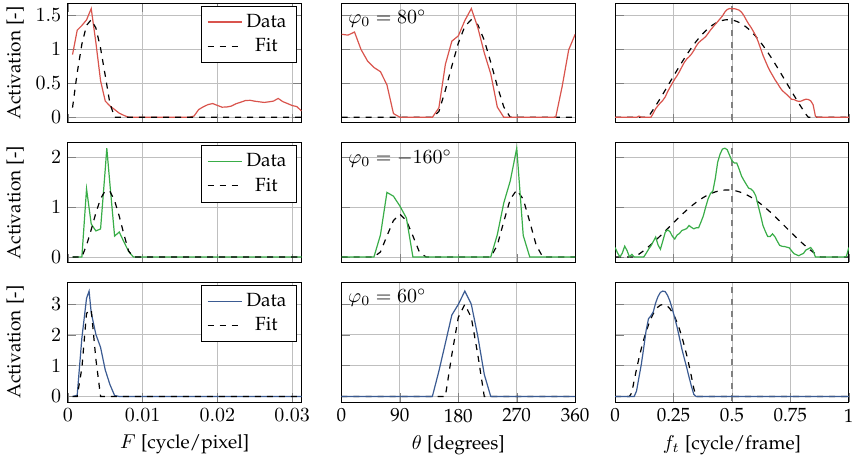}
	\end{subfigure}
	\caption{Quantitative results of the spectral Gabor filter fitting process.  \emph{Left:} Boxplot containing the total normalized cost $\mathcal{L}_{\text{norm}}$ per filter ($592$ filters). \emph{Right 3x3 plots:} Row-wise, the measured responses of three different \texttt{c6} filters and their corresponding Gabor fits. The blue, green, and red \texttt{c6} filters correspond to the crosses at the median, near the 75th percentile and near the upper whisker limit of the boxplot, respectively.}
	\label{fig:gaborfit}
\end{figure*}

\subsection{Results}

We found 592 of the 1024 filters in the \texttt{c6} layer of FlowNetS to have an activation larger than zero when using the aforementioned input waves. The location of the peak response of the active \texttt{c6} filters in terms of half spatial wavelength $\lambda_0 /2$, orientation $\theta_0$, and temporal frequency $f_{t_{0}}$ can be seen in Fig. \ref{fig:trans} (left). As shown, the locations of the peak responses of the filters are well distributed over all angles. Radially, there is a concentration around a half spatial wavelength of 200 pixels. Two possible explanations for this are the fact that (i) the average size of the chairs in the training dataset is 200 pixels, or that (ii) the half of the receptive field size of \texttt{c6} filters is 192 pixels. The concentration of the peak responses becomes even more apparent in Fig. \ref{fig:trans} (right), which shows the distribution along the temporal and half spatial wavelength axes. Furthermore, we note that the distribution of the temporal frequencies is skewed toward the Nyquist limit of $0.5$ cycles per frame. A possible reason for this is the low resolution in the temporal frequency due to the low number frames used as input to the network. This is further discussed in Section \ref{sec:tempband}. 

\begin{figure}[!t]
	\centering
	\includegraphics[width=0.95\linewidth]{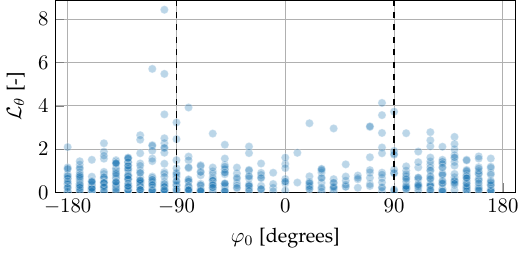}
	\caption{Orientation cost $\mathcal{L}_{\theta}$ per filter as a function of $\varphi_0$.}
	\label{fig:theta_error}
\end{figure}

\begin{figure*}[!t]
	\centering
	\begin{subfigure}{\linewidth}
	\centering
		\includegraphics[width=0.96\textwidth]{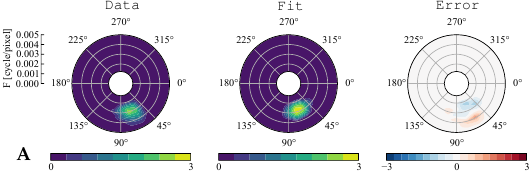}
	\end{subfigure}
	
	\begin{subfigure}{\linewidth}
	\centering
		\includegraphics[width=0.96\textwidth]{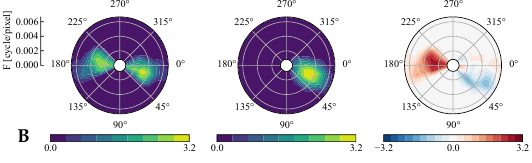}
	\end{subfigure}
	
	\begin{subfigure}{\linewidth}
	\centering
		\includegraphics[width=0.96\textwidth]{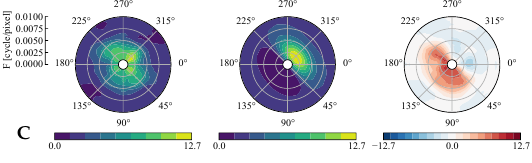}
	\end{subfigure}
	
	\begin{subfigure}{\linewidth}
	\centering
		\includegraphics[width=0.96\textwidth]{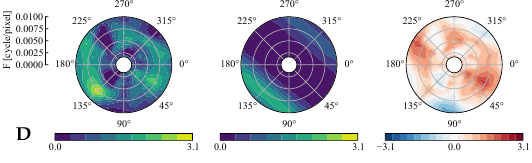}
	\end{subfigure}
	
	\caption[]{Qualitative results of the error patterns of the spectral Gabor fitting process. The spectral response profiles are shown as a function of spatial frequency $F$ and orientation $\theta$. \texttt{Data} shows to the measured response of a \texttt{c6} filter, \texttt{Fit} is the response of the corresponding fitted Gabor filter, and \texttt{Error} shows their difference.  Evaluations are with respect to $f_{t_{0}}$ and $\varphi_0$. (A) \texttt{c6} filter whose response profile is accurately captured by the Gabor model. (B) Red \texttt{c6} filter from Fig. \ref{fig:gaborfit}, which activates on opposite spatial frequencies. (C) \texttt{c6} filter with a very weak directional bias. (D) Noisy \texttt{c6} filter pattern (further discussed in Section \ref{sec:dilrot}).}
\end{figure*}

\begin{figure*}
	\ContinuedFloat
	\begin{subfigure}{\linewidth}
	\centering
		\includegraphics[width=0.96\textwidth]{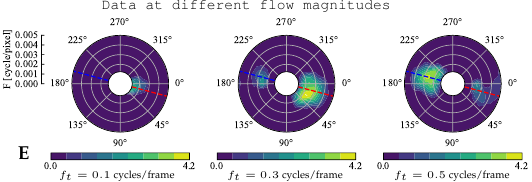}
	\end{subfigure}

	\caption{(continued) (E) For this \texttt{c6} filter, the spectral response profile for three different temporal frequency $f_t$ values is visualized. Two different Gaussian peak responses at opposite orientation can be observed at $f_t=0.3$ and $f_t=0.5$ cycles per frame. The blue and red lines correspond to the axes of the 2D representation of this filter shown in Figure \ref{fig:viz_occ}.}
	\label{fig:fit_viz}
\end{figure*}

\begin{figure}[!t]
	\centering
	\includegraphics[width=0.975\linewidth]{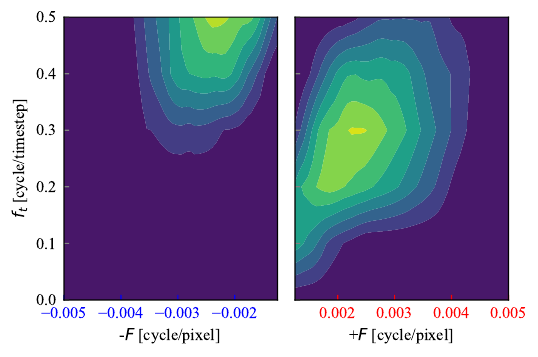}
	\caption{Spatiotemporal frequency representation of the measured filter response in Fig. \ref{fig:fit_viz}E. The positive and negative $F$-axes correspond to the blue and red lines in Fig. \ref{fig:fit_viz}E.}
	\label{fig:viz_occ}
\end{figure}

The main observation of our spectral analysis is that the fitted modified Gabor functions (i.e., Eq. \ref{eq:fitgab}) capture the spatiotemporal frequency selectivity of the active \texttt{c6} filters of FlowNetS accurately. In order to give insight into the goodness of fits for all neural responses in the \texttt{c6} layer, we show three example responses corresponding to different normalized cost values $\mathcal{L}_{norm}$ in Fig. \ref{fig:gaborfit}. Note that the fitted Gabor filters correspond well to the response of the blue and green \texttt{c6} filters (with $\mathcal{L}_{norm}$ at $50\%$, $75\%$ of the distribution); but, in the red case (an outlier), the fitted Gabor shows a substantial deviation from the measured \texttt{c6} response near $\theta=0$. 

\begin{table}[!t]
	\centering
	\caption{Result of the Gabor spectral response fitting procedure for different convolutional layers of the encoder part of FlowNetS.}
	\label{tab:layerfitting}
	\resizebox{.9\linewidth}{!}{
		\begin{tabular}{llll}
			\toprule
			\textbf{Layer}       & \textbf{$\mathcal{L}_{norm}$}  & \textbf{Max. $\lambda /2$} & \textbf{Num. active filters/filters} \\
			\midrule
			conv6\_1     & 1.65 & 800 & 592/1024                 \\
			conv5\_1     & 1.42 & 270 & 372/512                   \\
			conv4\_1     & 1.44  & 270& 408/512                   \\
 			conv3\_1     & 1.67  & 95 & 234/256                  \\
 			conv2        & 3.37 & 47  & 62/128                   \\
 			conv1        & 4.71 & 10  & 64/64                    \\
			\bottomrule                       
	\end{tabular}}
\end{table}

This experiment was also performed for the other convolutional layers of the network's encoder segment. As shown in Table \ref{tab:layerfitting}, the lower the layer, the smaller the receptive field size and hence the upper limit for the half spatial wavelength is decreased. According to the average (normalized) fitting error per layer \textbf{$\mathcal{L}_{norm}$}, the response of neurons in the \texttt{c3}--\texttt{c6} layers fits well the translational Gabor filter model, while our methodology suggests that neurons in \texttt{c1} and \texttt{c2} are not yet as motion-selective as Gabor filters. Table \ref{tab:layerfitting} also shows that \texttt{c6} is characterized by a higher \textbf{$\mathcal{L}_{norm}$} than its preceding layer. A possible explanation for this is that, in the earlier layers, the network is only able to perceive less complex motions which better fit the Gabor filter model.

\begin{figure}[!t]
	\centering
	\includegraphics[width=0.975\columnwidth, height=5.0cm]{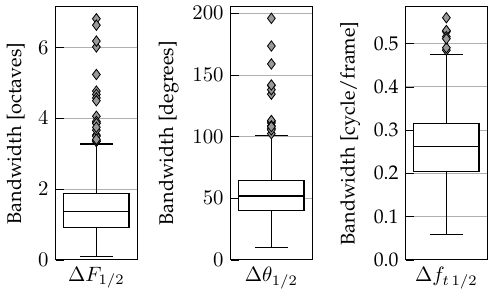}
	\caption{Bandwidth of spatial frequency $F$, orientation $\theta$, and temporal frequency $f_t$ of the fitted Gabor filters of the $75\%$ active \texttt{c6} filters with the lowest $\mathcal{L}_{norm}$.} 
	\label{fig:sigmas}
\end{figure}

Coming back to \texttt{c6}, the good fit for the majority of neurons supports the choice for the Gabor filter as opposed to other types of models. Of course, one can argue that the Gabor filter does not perfectly capture the response and a more complex model may lead to a better fit. Below, we will extensively delve into the cases in which the Gabor model seems to fall short of explaining \texttt{c6}'s neural responses. Here, it is important to note that in principle, we already have such a complex model: the neural network itself. The advantage of the Gabor model is that it has a low number of parameters that can be readily interpreted. Indeed, in neuropsychology, the step to more complex filters was only made when it became necessary for characterizing ``complex'' cells that did not respond to simple stimuli \cite{touryan2004nonlinear}. The fits and error patterns above the 75\% percent threshold (corresponding to the green \texttt{c6} filter) are very interesting, and we visually inspected them for systematic deviations. Visual inspection is performed instead of an auto-correlation procedure since the latter is not possible due to a non-uniformly spaced polar 3D frequency grid\cite{Jones1987}. Fig. \ref{fig:fit_viz} contains the qualitative results used for this analysis, while Appendix C evaluates the generalizability of the fitted Gabor filters to more complex natural stimuli. 

Similarly to the blue filter in Fig. \ref{fig:gaborfit}, Fig. \ref{fig:fit_viz}A shows a \texttt{c6} filter whose response fits nicely in the Gabor filter framework. On the other hand, we find three types of systematic deviations (i.e., Fig. \ref{fig:fit_viz}B, \ref{fig:fit_viz}C, \ref{fig:fit_viz}E) from the Gabor model, and also conclude that some patterns are too complex for interpretation, such as the \texttt{c6} filter shown in Fig. \ref{fig:fit_viz}D.

The filter in Fig. \ref{fig:fit_viz}B shows a deviation from the fitted Gabor 180 degrees away from $\theta_0$. This filter is responsive to edge structure (i.e., $|\varphi_0| \approx 90^{\circ}$) and is thus approximately odd, since the dot product of two odd signals at opposite frequencies results in a negative value. However, this filter still produces a positive activation at the opposite spatial frequency, corresponding to 180 degrees away from $\theta_0$. In Fig. \ref{fig:theta_error} the distribution of the phase values $\varphi_0$ versus orientation cost $\mathcal{L}_{\theta}$ for all filters is depicted. As shown, there are multiple filters responsive to edge structure that have a high $\mathcal{L}_{\theta}$ (e.g., the red filter in Fig. \ref{fig:gaborfit}). One possible reason for this systematic deviation from the Gabor response is that the network is able to learn flow filters that are invariant to polarity (meaning white-black or black-white transitions).

We find two \texttt{c6} filters that exhibit weak directional bias, an example of which can be found in Fig. \ref{fig:fit_viz}C. Moreover, we also find filters that exhibit two or more Gaussian peaks with similar peak response magnitudes but tuned to different spatial frequencies $F_0$, orientations $\theta_0$, and temporal frequencies $f_{t_{0}}$. An example of such a filter can be found in Fig. \ref{fig:fit_viz}E, and its 2D spatiotemporal representation is shown in Fig. \ref{fig:viz_occ}. A possible explanation is that these filters are sensitive to occlusion, as discussed in Section \ref{sec:dilrot}. Lastly, we find filters that appear noisy and are hard to interpret given the limitations of our methodology (further discussed in Section \ref{sec:dilrot}). Such an example can be seen in Fig. \ref{fig:fit_viz}D.

\subsection{Temporal bandwidth}
\label{sec:tempband}

For orientation $\theta$ and temporal frequency $f_t$, the bandwidth is defined as the width of the filter which provides an output above half the maximum response. This leads to a bandwidth in degrees $\Delta \theta_{1/2}$ and cycles per frame $\Delta f_{t_{1/2}}$ for orientation and temporal frequency respectively:
\begin{equation}
\Delta f_{t_{1 / 2}}=f_{t_{\max}}-f_{t_{\min}}
\end{equation}
\begin{equation}
\Delta \theta_{1 / 2}=\theta_{\max }-\theta_{\min }
\end{equation}

For spatial frequency $F$, the bandwidth is defined in terms of octaves as follows: 
\begin{equation}
\Delta F_{1 / 2}=\log _{2}\left(F_{\max } / F_{\min }\right)
\end{equation}

\setcounter{figure}{9}
\begin{figure*}[!b]
	\centering
	\includegraphics[width=1.\linewidth]{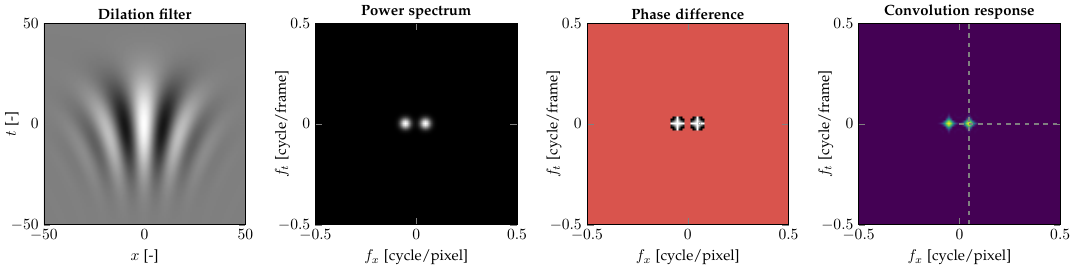}
	\caption{Convolution response of a dilation filter $dw$ with a translating plane wave $s$ evaluated with spatiotemporal frequencies at $k$ integer multiples of the fundamental frequency. In the $\psi$ plot, a larger phase difference corresponds to a darker color with black being equal to or greater than $\pi/2$. A red mask is applied to frequency components with low power. The dashed lines indicate the Gaussian pattern perceived by the spectral fitting procedure.}
	\label{fig:dil}
\end{figure*}   

\setcounter{figure}{8}
\begin{figure}[!t]
	\centering
	\includegraphics[width=.9\columnwidth]{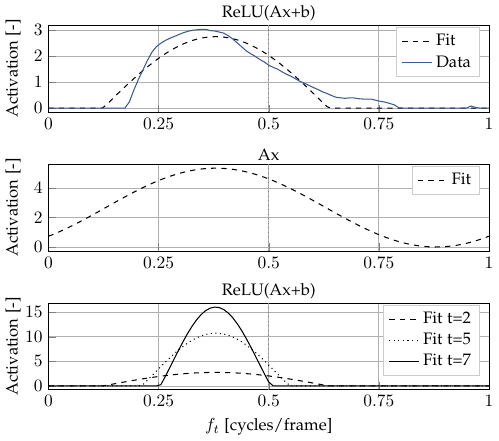}
	\caption{Illustration of how the network is able to decrease the extent of the filter response in the temporal domain. \emph{Top:} Fit and measured data for the median \texttt{c6} filter (see Fig. \ref{fig:gaborfit}). \emph{Middle:} The response of the fitted Gabor filter without the bias term and ReLU non-linearity. \emph{Bottom:} Response of the fitted Gabor filter when the number of frames is increased.
	} 
	\label{fig:timsam}
\end{figure}

Although we estimate the Gabor parameters of the active \texttt{c6} filters in the fitting process, the apparent bandwidth of these filter differs due to the non-linear transform in Eq. \ref{eq:fitgab}. The bandwidth is therefore measured based on the fitted Gabor filter response. In Fig. \ref{fig:sigmas}, the bandwidth of $F$, $\theta$, and $f_t$ can be seen. As shown, the interquartile range for spatial frequency bandwidth is between 1 and 2 octaves and the median orientation bandwidth is approximately $50^{\circ}$. Lastly, the temporal frequency bandwidth is of large extent with a median of approximately $0.27$ cycles per frame. 

We note that the network is able to narrow the extent of the filter response in the temporal domain using the non-linear transform in Eq. \ref{eq:fitgab}. An illustration of this mechanism can be seen in Fig. \ref{fig:timsam}. As shown, the extent of the half-magnitude profile is wider if the non-linear transformation is not employed. This figure also shows what happens when more frames are added to the input and the other parameters are kept the same (see Fig. \ref{fig:timsam}, bottom). This suggests that an even narrower extent could be reached by feeding the network with more images over time than just the two subsequent images used in FlowNetS. A higher resolution in the frequency domain is beneficial as it allows for a more precise measurement of the flow.

\setcounter{figure}{10}
\begin{figure*}[!t]
	\centering
	\includegraphics[width=1.\linewidth]{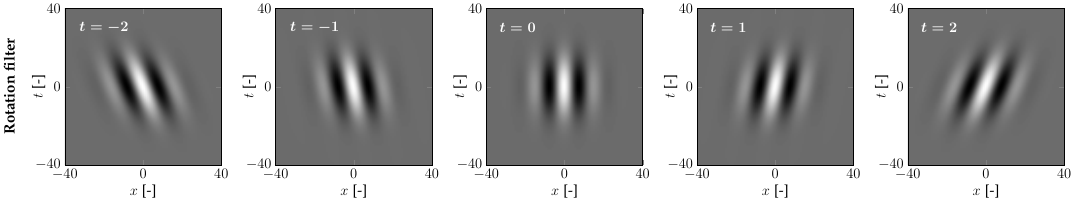}
	\includegraphics[width=1.\linewidth]{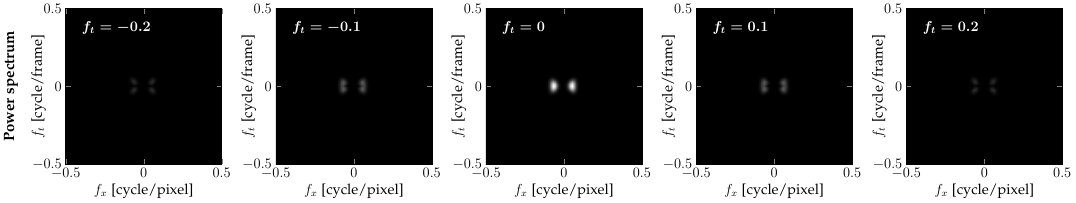}
	\includegraphics[width=1.\linewidth]{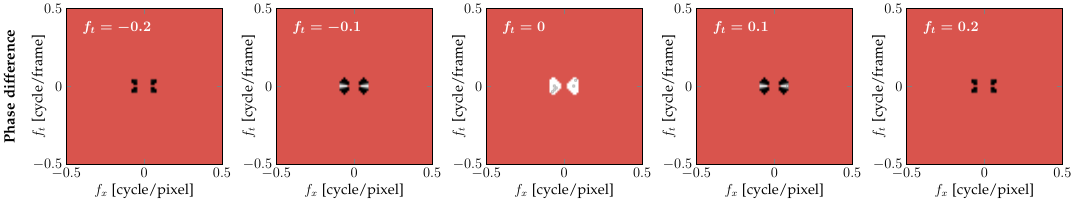}
	\includegraphics[width=1.\linewidth]{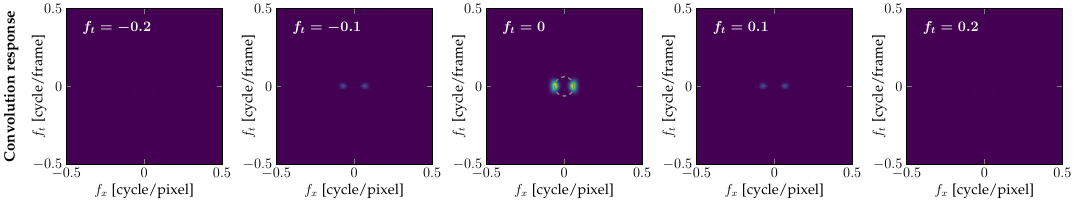}
	\caption{Convolution response of a rotation filter $cw$ with a translating plane wave $s$ evaluated with spatiotemporal frequencies at $k$ integer multiples of the fundamental frequency. In the $\psi$ plot, a larger phase difference corresponds to a darker color with black being equal to or greater than $\pi/2$. A red mask is applied to frequency components with low power. The dashed circle indicates the double lobe Gaussian pattern perceived by the spectral fitting procedure.}
	\label{fig:rot}
\end{figure*}

\section{Network response to dilation \& rotation}\label{sec:dilrot}

In this section, the sensitivity of \texttt{c6} filters to dilation and rotation is analyzed. First, we explain the limitations of the spectral Gabor response profile fitting process and why we are not able to discern filters activating on translation, dilation, rotation, and occlusion with this methodology. Second, the theory used to identify filters sensitive to dilation and rotation is presented. Lastly, our results are discussed.

Note that Gabor translation filters \cite{Gabor1945} and occlusion filters \cite{Beauchemin2000} already have an analytical description in both the space-time and frequency domain. Such a description of dilation and rotation is, to the best of the authors' knowledge, missing. Therefore, fitting \texttt{c6} filters to a dilation and rotation motion filter model requires a novel mathematical foundation which is outside of the scope of this work.

\subsection{Limitations of the spectral response profile fitting}

In the first part of the spectral response fitting process, a gridsearch is performed to find the peak response. In the subsequent fitting process, three response lines are generated by varying either $F$, $f_{t}$, or $\theta$, whilst keeping $\varphi$ constant. This method only allows the measurement of the relative attenuation in amplitude with respect to the peak response $\hat{r}_0$. This is sufficient for translation, which can be defined as a single constant phase Gaussian in the 3D frequency spectrum and thus produces a Gaussian in response. However, it is insufficient for other more complex motion types.

Due to the ReLU activation function, the dot product of two translating plane waves at the same frequency, which are more than or equal to 90 degrees out-of-phase, is zero. Note that a convolution in the space-time domain equals to multiplication in the frequency domain according to the convolution theorem\cite{Bracewell1986}. Because we evaluate the convolution response only at discrete frequencies of $k$ integer multiples along the $f_x$, $f_y$, and $f_t$ axis, only a single frequency component of the Fourier-transformed translating plane wave $S$ will contain power\footnote{Not taking into account the complex conjugate component.}. Then, if we define the $k$-th frequency component of $S$ as the complex vector $\mathbf{p}$, and the $k$-th frequency component of the Fourier transformation of the filter to be analyzed as $\mathbf{q}$, the phase difference between these two complex vectors is defined as the angle $\psi$ and given by:
\begin{equation}
\psi = \cos^{-1}( \frac{\mathbf{p} \cdot \mathbf{q}}{|\mathbf{p}| |\mathbf{q}|} )
\end{equation}
\noindent where the maximum value of $\psi$ is $\pi$, and values of $\psi\geq\pi/2$ result in a zero response due to the ReLU in Eq. \ref{eq:fitgab}.

\begin{figure*}[!t]
	\centering
	\includegraphics[width=1.\linewidth]{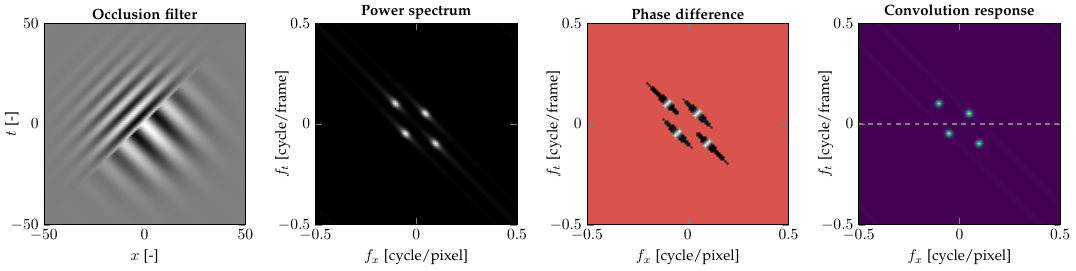}
	\caption{Convolution response of an occlusion filter with a translating plane wave $s$ evaluated with spatiotemporal frequencies at $k$ integer multiples of the fundamental frequency. \emph{Left:} Example occlusion signal following the description of \textit{Beauchemin et al.}\cite{Beauchemin2000}. \emph{Middle left:} The power spectrum of the Fourier-transformed occlusion filter. \emph{Middle right:} The angle $\psi$ indicating the phase difference between the Fourier components of the occlusion filter and $s$. A larger phase difference corresponds to a darker color with black being equal to or greater than $\pi /2$. A red mask is applied to frequency components with low power. \emph{Right:} Convolution response between the occlusion filter and $s$. The pattern above the dashed gray line resembles that of Figs. \ref{fig:fit_viz}E and \ref{fig:viz_occ}.}
	\label{fig:occ_sim}
\end{figure*}

\noindent \textbf{Convolution response: Dilation \& rotation filters}\\
\noindent To determine which frequency components of dilation, rotation, and occlusion are more than 90 degrees out of phase, the Discrete Fourier Transform (DFT) \cite{Bracewell1986} is used to transform a simulated space-time signal to a representation in the frequency domain. Fig. \ref{fig:dil} shows the convolution response of a dilation filter $dw$ with a translating plane wave $s$. From this figure, it can be observed that a diamond-like pattern emerges in the response, due to the immeasurable out-of-phase components of $dw$ and $s$. Because we evaluate the responses along lines orthogonal to the peak response, the pattern perceived is indicated by the dashed lines in the right-most plot of this figure, which correspond to the colored linear patterns in Fig. \ref{fig:gaborspace}. Thus, a Gaussian will be perceived along the spatial and the temporal frequency ranges. Hence, we are not able to discern between dilation and translation filters. 

Similarly, Fig. \ref{fig:rot} shows the convolution response of a rotation filter $cw$ with $s$. Note that the 3D power spectrum of $cw$ is different from a Gaussian. At high temporal frequencies (i.e., $\pm0.2$ cycles per frame), the frequency components of $cw$ and $s$ are out-of-phase. Thus, these frequency components will not be detected. The pattern perceived along the varying $\theta$ (also shown in Fig. \ref{fig:gaborspace}) is two Gaussian lobes at opposite frequency. This pattern is similar to the convolution response of a cosine Gabor filter tuned to stationary patterns (i.e., zero temporal frequency). Therefore, our methodology is also not able to detect rotation filters.

\noindent \textbf{Convolution response: Occlusion filters} \\
\noindent Furthermore, we convolve an occlusion filter, using the description of \textit{Beauchemin et al.} \cite{Beauchemin2000}, with translating plane waves $s$. Occlusion in the spatiotemporal domain can be described as the combination of a Gaussian, a Heaviside step function, and two translating plane waves translating with different frequencies, as shown in Fig. \ref{fig:occ_sim}. The power spectrum of the Fourier-transformed filter can be described as two Gaussian filter pairs with \textit{tails} due to the Heaviside step function. The angle $\psi$ demonstrates that these tails have a large phase difference. Consequently, only the pattern above the dashed line is detected using our methodology, which corresponds to two different Gaussian lobes tuned to different frequencies. This pattern resembles that of Figs. \ref{fig:fit_viz}E and \ref{fig:viz_occ}, thus making it likely that the filter represented in these figures is responsive to occlusion. However, it should be noted that we are not able to discern such a pattern from the superposition of two regular Gabor filter pairs tuned to different frequencies.

\begin{figure*}[!t]
	\centering
	\begin{subfigure}[t]{\textwidth}
	\begin{subfigure}[t]{0.5\textwidth}
		\centering
		\includegraphics[width=0.8\linewidth]{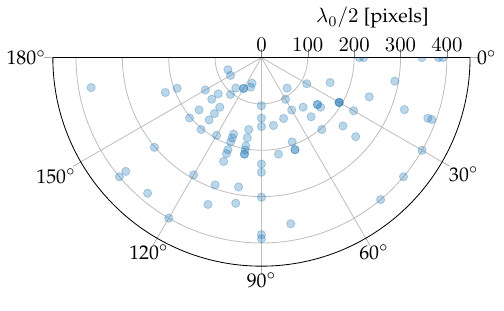}
	\end{subfigure}%
	~ 
	\begin{subfigure}[t]{0.5\textwidth}
		\centering
		\includegraphics[width=0.775\linewidth]{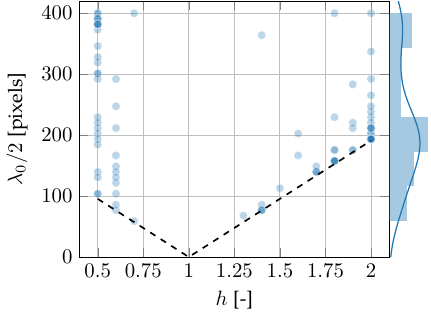}
	\end{subfigure}
	\caption{\emph{Left:} Half spatial wavelength $\lambda_0/2$ and initial orientation $\theta_0$. \emph{Right:} Half spatial wavelength $\lambda_0/2$ and scale factor $h$. The black dashed line indicates the temporal aliasing constraint given by Eq. \ref{eq:dilconstr}.}
	\label{fig:grid_dil}
	\end{subfigure}\\\vspace{10pt}
	\begin{subfigure}[t]{\textwidth}
	\begin{subfigure}[t]{0.5\textwidth}
		\centering
		\includegraphics[width=0.8\linewidth]{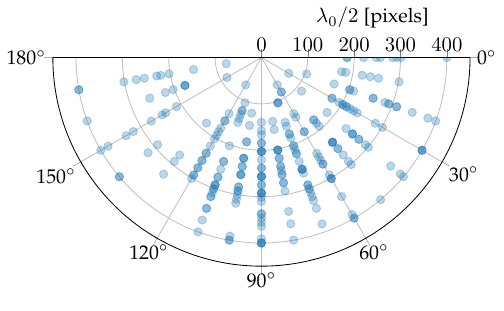}
	\end{subfigure}%
	~ 
	\begin{subfigure}[t]{0.5\textwidth}
		\centering
		\includegraphics[width=0.775\linewidth]{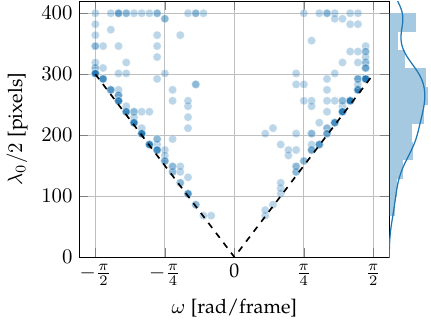}
	\end{subfigure}
	\caption{\emph{Left:} Half spatial wavelength $\lambda_0/2$ and initial orientation $\theta_0$. \emph{Right:} Half spatial wavelength $\lambda_0/2$ and angular temporal frequency $\omega$. The black dashed line indicates the temporal aliasing constraint given by Eq. \ref{eq:rotconstr}.}
	\label{fig:grid_rot}
	\end{subfigure}
	\caption{Location of peak response $\hat{r}_0$ per \texttt{c6} filter in the spatiotemporal frequency domain in response to dilating (top) and rotating waves (bottom). Only filters whose peak response $\hat{r}_0$ was higher than the maximum found in the translation gridsearch are shown.}
\end{figure*}

\subsection{Methodology}

In order to still assess the sensitivity of the \texttt{c6} filters to dilation and rotation, we come up with a different methodology in which two gridsearches are performed. We assess the locations of the peak responses for filters which have a higher response to dilation or rotation than to translation. We do not classify a filter as either a rotation or dilation filter, since a filter can be sensitive to a composition of these respective motions.
\\ \noindent \textbf{Dilation parametrization} \\
\noindent As in \cite{Fleet1992}, a dilating wave $d$ is given by:
\begin{equation}
d(x, y,t)=\cos \left(2 \pi F_{0} (x_r-\alpha x_r t) +\varphi_{0}\right)
\label{eq:dilat}
\end{equation} 
\noindent where $\alpha$ denotes the dilation factor. The training dataset used to train FlowNetS, i.e. FlyingChairs \cite{Dosovitskiy2015}, defines scaling motion in terms of the affine scaling factor $h$. Because the network only takes two frames as input, we define the relation between $h$ and $\alpha$ as follows:
\begin{equation}
h = \frac{1}{1-\alpha}
\end{equation}

The gridsearch is performed for the $[0.5,2.0]$ range of $h$, as it encapsulates the values encountered during training. More details about this search space can be found in Appendix B. In order to mitigate the effect of temporal aliasing, the search space is constrained so that the velocity of a point is not more than half its spatial wavelength $\lambda_0/2$. For a dilating wave, this velocity is given by:
\begin{equation}
v = \big(\frac{1}{1-\alpha}-1\big)x = (h-1)x
\end{equation}

Then, the temporal aliasing constraint for dilating waves is given by: 
\begin{equation}
(h-1)x \leq \frac{1}{2} \lambda_0
\label{eq:dilconstr}
\end{equation}

\noindent \textbf{Rotation parametrization} \\
\noindent A rotation wave $c$ is given by:
\begin{equation}
c(x, y,t)=\cos \left(2 \pi F_{0} x_r(t)+\varphi_{0}\right)
\label{eq:rotation}
\end{equation} 
\noindent where $x_r(t)$ varies with time, and is defined as:
\begin{equation}
x_r(t)=x \cos (\theta_0 + \omega t) + y \sin (\theta_0 +\omega t)
\end{equation}
\noindent where $\omega$ denotes the angular velocity in radians per frame. 

The search space for the rotation gridsearch can be found in Appendix B. A constraint was also added to limit the effect of temporal aliasing. $\omega$ can be related to a point at distance $m$ from the center of rotation by $v=\omega m$. The maximum distance from the center of rotation to the edge is equal to half the receptive field size, which is 383 pixels in the \texttt{c6} layer of our FlowNetS. As the wave rotates around the center pixel, the velocity at this point should thus be lower than half the spatial wavelength. The constraint is given by the following relation:
\begin{equation}
\omega m_{\text{max}} \leq \frac{1}{2} \lambda_0
\label{eq:rotconstr}
\end{equation}

\subsection{Results}
The peak responses of \texttt{c6} filters which have a higher activation to dilation than to translation (i.e., approximately $15\%$ of the active filters) are shown in Fig. \ref{fig:grid_dil}. These filters show a radially dispersed pattern along the $\theta$-axis, and a peak in the distribution of half spatial wavelengths near 200 pixels. Lastly, peak responses are often close to the temporal aliasing limit and the maximum scaling value of the gridsearch. This is similar to the temporal peak response location for the translation gridsearch (see Fig. \ref{fig:trans}).

In Fig. \ref{fig:grid_rot}, the peak responses of the \texttt{c6} filters for the rotation gridsearch are shown. It can be observed that most filters are active near the temporal translation and temporal rotational aliasing limit. Also, a peak in the distribution of half spatial wavelengths can be identified around 250 pixels, which is slightly higher than expected. A possible explanation for this discrepancy is that rotation is actually a 3D motion and thus the scale should also be limited along its radial axis. Approximately $45\%$ of the active \texttt{c6} filters activate more on rotation than on translation, which could be due to the fact that we do not limit the wavelength along the axis of rotation. The points in the motion field at the far end of the receptive field then move with a very high velocity, and therefore, the response of the filters is higher.

\begin{figure*}[!t]
	\centering
	\begin{subfigure}[c]{0.5\textwidth}
		\centering
		\includegraphics[width=1\linewidth]{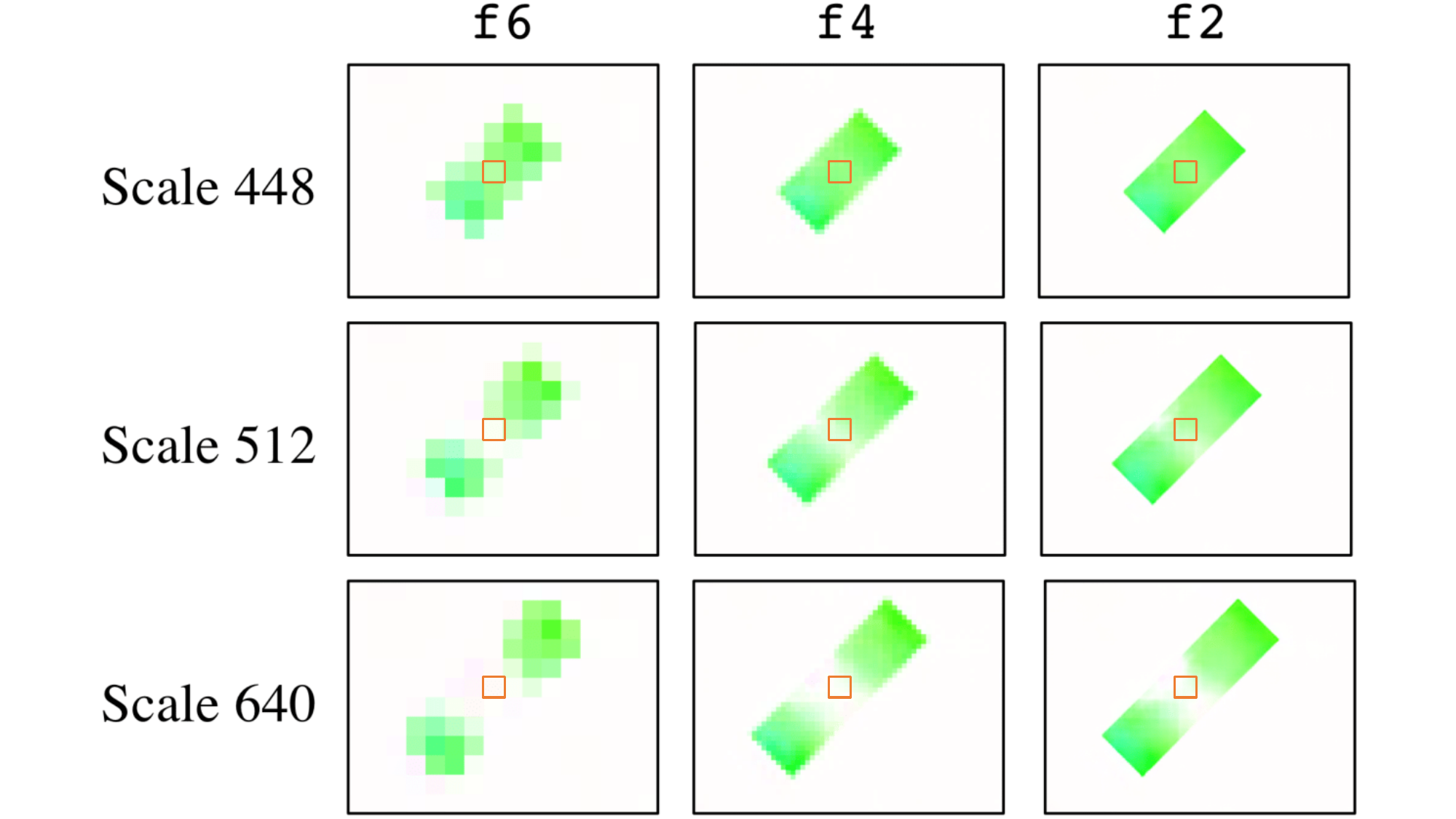}
		\label{fig:apfmap}
	\end{subfigure}%
	~
	\begin{subfigure}[c]{0.5\textwidth}
		\centering
		\includegraphics[width=0.95\linewidth]{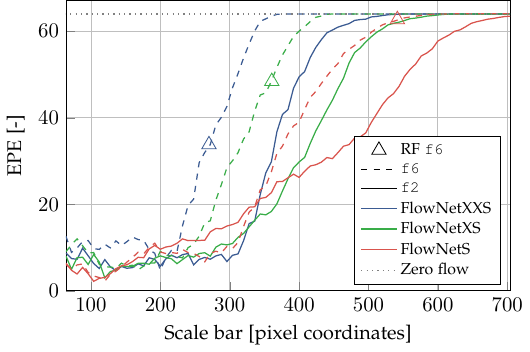}
		\label{fig:aperror}
	\end{subfigure}
	\caption{Response of our FlowNetS and its two variations, FlowNetXS and FlowNetXXS, to diagonally translating bars with motion magnitude $|\mathbf{u}|=64$ pixels. \textit{Left:} \texttt{f6}, \texttt{f4} and \texttt{f2} FlowNetS flow maps in response to downward-left diagonally translating bars of different scales, using the color-coding scheme from \cite{Baker2011}. The red squares highlight the output region used for evaluating the error. \textit{Right:} EPE versus scale of the bar in pixel coordinates. RF \texttt{f6} indicates the diagonal receptive field size in pixel coordinates corresponding to the \texttt{f6} flow map.
	}
	\label{fig:aperture}
\end{figure*}

\section{Solving the aperture problem}
\label{sec:ap}

\subsection{Methodology}

In order to determine until what scale of input stimuli FlowNetS can resolve the aperture problem, three different versions of this network are trained under the same circumstances with varying receptive field sizes. The receptive field size is defined as the region in the input images which affects the value of the feature map at a particular layer and feature map location. Therefore, we modify the filter size of the convolutional kernels in \texttt{c6}, which is actually composed of two layers: \texttt{c6\_0} and \texttt{c6\_1}. The original (and our) FlowNetS uses 3x3 kernels in these layers, which leads to a receptive field size of 383 pixels in the \texttt{f6} flow map. We train two additional models with kernels sizes (1x1, 3x3) and (1x1, 1x1) for \texttt{c6\_0} and \texttt{c6\_1}, which we name FlowNetXS and FlowNetXXS, and whose \texttt{f6} receptive field size is 255 pixels and 191 pixels, respectively. For the three of these networks, the receptive field size increases in the expanding part of the architecture due to the upconvolutional layers.

As input, we use a diagonally translating bar of different scales with motion magnitude $|\mathbf{u}|=64$ pixels. We determine the error at the center of the bar, and at three flow maps of different resolutions: \texttt{f6}, \texttt{f4}, and \texttt{f2} (see Fig. \ref{fig:archflownet}).

\subsection{Results}

In Fig. \ref{fig:aperture} (left), the FlowNetS response to a downward left translating bar of varying scale is shown. Firstly, the flow becomes more and more refined in the expanding part of the architecture. Secondly, the network is able to extrapolate motion cues from the edges of the bar towards the center, but only to an extent determined by the scale of the bar.

Fig. \ref{fig:aperture} (right) shows the average End-Point-Error (EPE) of FlowNetS, FlowNetXS, and FlowNetXXS in response to two translating bars of different scales moving upward right and downward left, respectively. As shown, the network's robustness to the aperture problem is related to the receptive field size, and networks with larger receptive fields are able to resolve the aperture problem at larger scales. 

\section{Discussion and future work}
\label{sec:dis}

\subsection{Impact on Computer Vision}
Our results show that the neural responses in the deepest encoding layer of FlowNetS, \texttt{c6}, are well captured by Gabor-like filters. This finding provides insight into the limits and robustness of the approach. Given this core mechanism for estimating optical flow, it is to be expected that the network generalizes quite well to out-of-training-set samples. However, it also raises some concerns, since traditional Gabor filters for optical flow estimation had certain disadvantages. They deal badly with deviations from translation, varying contrast due to changing lighting conditions, and are subject to the uncertainty relation, which corresponds to the balance between localization of the stimuli in the spatial domain and resolution in the frequency domain. 

FlowNetS successfully copes with all of these issues. We have shown that deviations from translations are dealt with by additional filters that are sensitive to more complex motion types. Moreover, \textit{Mayer et al.} \cite{Mayer2018} showed that FlowNet is able to cope with varying contrast over time due to changing lighting conditions. Lastly, we have demonstrated that FlowNetS is able to achieve a better spatial localization of motion cues in the expanding part of the network, thus coping with the uncertainty relation.

In terms of accuracy, FlowNetS did not reach the levels of state-of-the-art methods. For example, it has poor performance on sub-pixel flow \cite{Ilg2017}. One reason for this might be the large number of strides utilized before the initial flow prediction is made. Also, our analysis shows that a Gabor filter based on two frames results in a large temporal frequency bandwidth, and hence limited performance concerning flow velocity estimation. This is narrowed somewhat by the non-linear transformations due to the ReLU activation function and bias term. However, our analysis indicates that this could be further improved by using more frames and thus providing more temporal information to the network. Please note that there is an increasing number of multi-frame methods for deep optical flow estimation, e.g.,  \cite{neoral2018continual,liu2019selflow,guan2019unsupervised,godet2020starflow}. As remarked in \cite{neoral2018continual}, most of these methods use multiple images in order to track flow to future frames and track flow back to the past, in order to enhance consistency of the flow. Methods such as StarFlow \cite{godet2020starflow} additionally pass the flow and extracted features from the previous image pairs as input to the deep net, while other methods make use of LSTMs \cite{guan2019unsupervised}. However, the basic matching still happens between two frames with FlowNetC-like neural correlation blocks. What we propose here is to enter multiple images directly into a FlowNetS-like network in order to reduce the temporal bandwidth, something which to our knowledge has not been investigated yet. 

The Gabor-like nature of the neural filters in \texttt{c6} may also be a reason for less accuracy; These responses are mapped to coarse flow in a linear way by \texttt{pf6}. This means that optical flow velocities that are higher than the filter's tuned velocity, actually lead to an underestimation of the optical flow (due to the bell-shape of the response, see, e.g., Fig. \ref{fig:gaborfit}). The network likely copes with this in the following ways. First, it can narrow the response bandwidth with the nonlinear activation function. To see why this helps, think of the extreme in which a neuron would respond in a Dirac-like way to a very specific optical flow velocity. Of course, such a narrow response would then require a very large number of neurons to cover all optical flow velocities. This brings us to the second coping mechanism; The final flow is mostly determined by the neurons in the neural filter bank that are tuned closer to the true optical flow velocity, as they will react more intensely. Finally, the biases in the network can be set in a way to deal with this problem, which is biased since it mostly involves underestimation. Still, it may be worth investigating if different mechanisms would lead to a better accuracy, for instance by introducing a winner-take-all mechanism. 

We observed that only 592 of the 1024 \texttt{c6} filters have an activation larger than zero. However, the high similarity of the active filters to the Gabor model already suggests that it would also be worth studying a hybrid FlowNetS network, in which there is a fixed Gabor filter bank (extended with rotation and dilation features) followed by a convolutional multi-layer loss flow refinement. This would greatly reduce training time, and, most probably, improve the generalizability of the network.

Finally, our findings for FlowNetS may also be relevant to ``PoseNets'' (e.g., \cite{zhou2017unsupervised,chen2019self}) that take as input subsequent images and output the relative pose, i.e., an estimate of the translation and rotation between them. Typically, for such relative pose estimation networks a simple encoder structure is used, which is very similar to FlowNetS's structure up to and including \texttt{c6}. We expect that optical flow plays a large role in the estimation of translation and rotation between subsequent images, and - given the similar network structure - it is possible that PoseNets also implicitly determine flow with Gabor-like filters before synthesizing the information into a translation and rotation estimate.

\subsection{Impact on biology}
We have used and extended methods from neuropsychology for determining the types of motion filters represented by neurons in the deep \texttt{c6} layer of FlowNetS. The analysis gave very similar results to those on neurons in the mammalian visual cortex. First, many filter responses fit very accurately with Gabor filters that capture translational motion. Second, the spatial and orientation bandwidth statistics show similarity to bandwidths of neurons found in the mammalian visual cortex. We report a median spatial frequency bandwidth of 1.36 octaves, while \textit{De Valois et al.} \cite{DeValois1982a} report 1.4 octaves for the macaque visual cortex. Similarly, we find a median orientation bandwidth of 52 degrees, while \textit{De Valois et al.} \cite{DeValois1982b} find 65 degrees. These similarities may be due to similar optical flow statistics being perceived both by the network and the animals. Third, as in neuropsychological experiments \cite{Deangelis1993}, we observed that some filters respond poorly to translating plane waves. Our analysis shows that such poor response may be due to the filters being sensitive to more complex motions such as dilation and rotation. Indeed, in the human brain, channels sensitive to dilation have been found \cite{Regan1978a}. However, this did not provide conclusive evidence of neurons sensitive to dilation. Our analysis and results suggest that it is worth looking for dilation- and rotation-sensitive neurons in animal brains. In fact, one could even extend the analysis to also check for shear, as this forms an additional basis for the flow field derivatives \cite{Longuet-Higgins1980}. 

\section{Conclusion}
\label{sec:conc}
We have employed a spectral response fitting approach from neuropsychology to demonstrate that the deepest layer of FlowNetS essentially encodes a bank of spatiotemporal Gabor filters. Although accurate fits were obtained, the spectral response fitting approach is limited, since it is not able to identify the exact motion pattern causing the maximum activation of a filter. In this work, we have already shown that the network also contains a large number of filters that are more sensitive to dilation and rotation than to translation, but more complex motion filters may be present. Finally, we have studied how FlowNetS tackles the aperture problem. Our results suggest that, on the one hand, the receptive field size is highly correlated to the scale at which the network can resolve the aperture problem. On the other hand, the expanding part of the network allows to solve the aperture problem at slightly larger scales by performing a filling-in function similar to that in mammal vision systems.

Future work could: (i) perform a similar analysis on SpyNet \cite{Ranjan2017a}, (ii) study the neural response to more complex motion patterns like compositions of affine and 3D motion, as present in more realistic synthetic training datasets (e.g., FlyingThings \cite{Mayer2016}), (iii) attempt to improve FlowNetS' performance by using smaller strides or more input images, and (iv) employ our extended spectral response fitting method to investigate if animal brains have dilation- and rotation-sensitive neurons as well.

\bibliographystyle{IEEEtran}
%\bibliography{IEEEabrv,refs}

\begin{IEEEbiography}[{\includegraphics[width=1in,height=1.25in,clip,keepaspectratio]{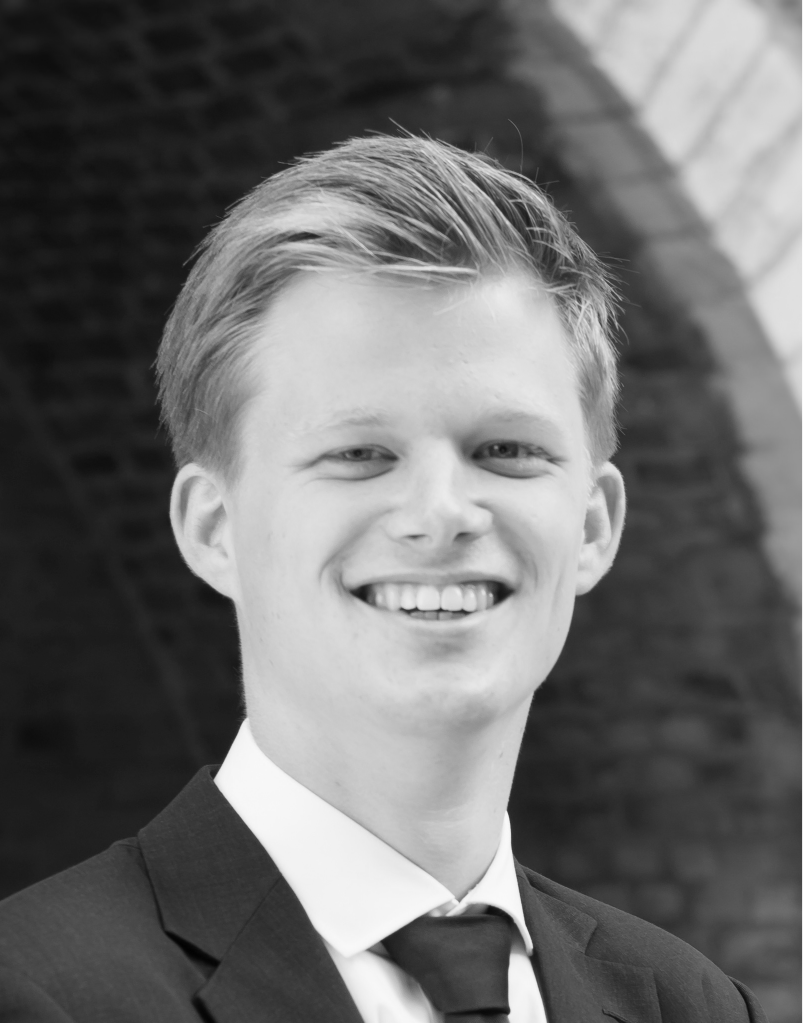}}]{David B. de Jong}
	received his B.Sc. in Aerospace Engineering from the Delft University of Technology, the Netherlands, in 2016, and his M.Sc in Aerospace Engineering at the Control \& Simulation department from the same university in 2020. His research interest is the intersection of machine learning, neuroscience, and computer vision.
\end{IEEEbiography}

\begin{IEEEbiography}[{\includegraphics[width=1in,height=1.25in,clip,keepaspectratio]{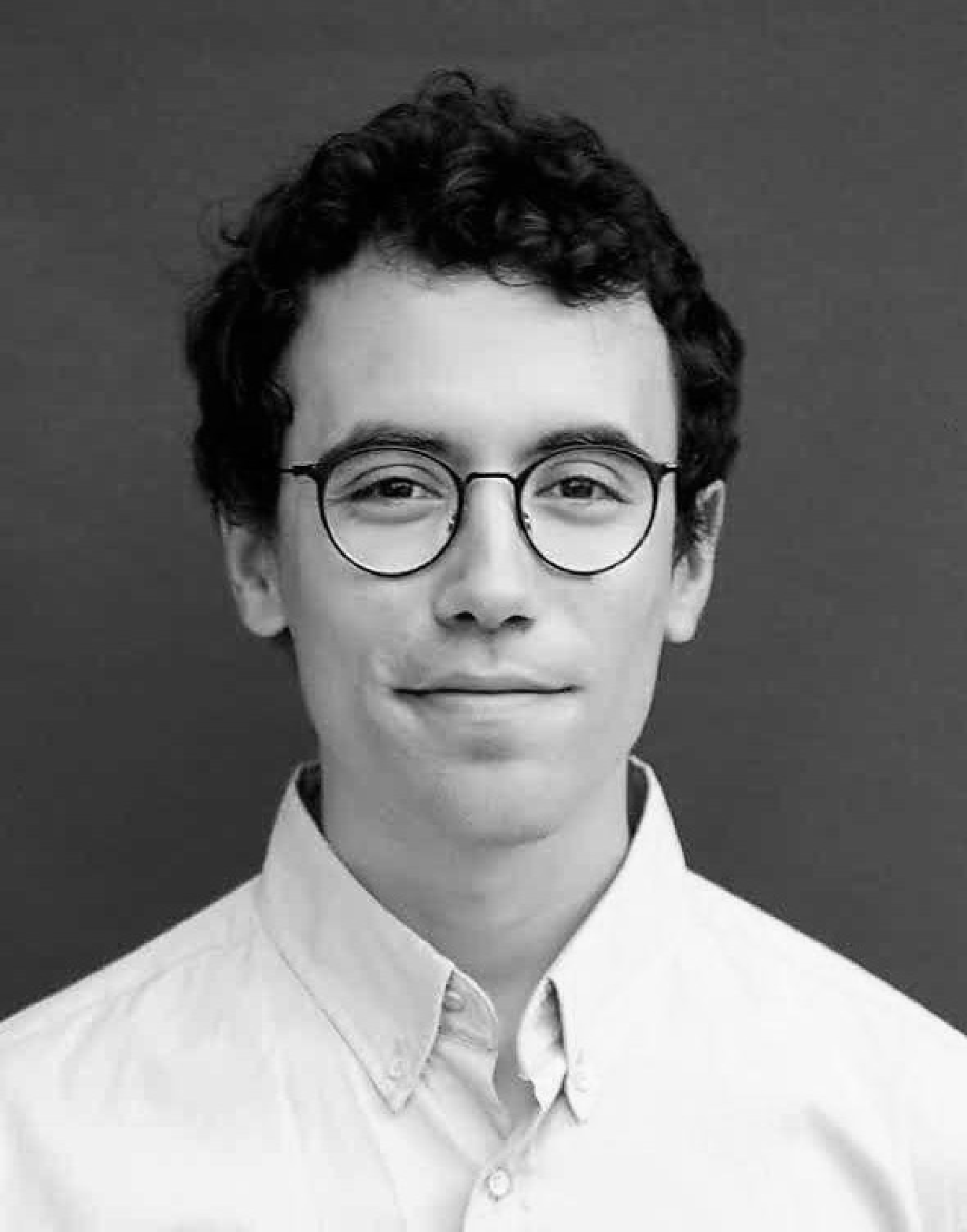}}]{Federico Paredes-Vall\'es}
	received his B.Sc. in Aerospace Engineering from the Polytechnic University of Valencia, Spain, in 2015, and his M.Sc. from Delft University of Technology, the Netherlands, in 2018. He is currently a Ph.D. candidate in the Micro Air Vehicle Laboratory at the latter university. His research interest is the intersection of machine learning, neuroscience, computer vision, and robotics.
\end{IEEEbiography}

\begin{IEEEbiography}[{\includegraphics[width=1in,height=1.25in,clip,keepaspectratio]{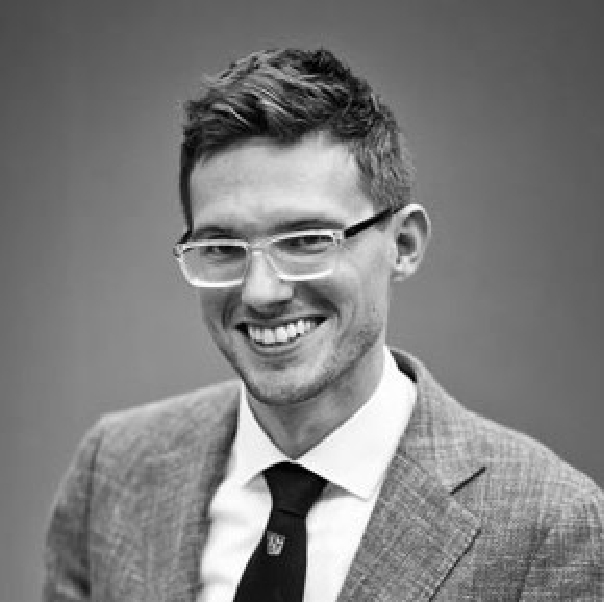}}]{Guido~C.~H.~E.~de~Croon}
	received his M.Sc. and Ph.D. in the field of Artificial Intelligence at Maastricht University, the Netherlands. His research interest lies with computationally efficient algorithms for robot autonomy, with an emphasis on computer vision. Since 2008 he has worked on algorithms for achieving autonomous flight with small and light-weight flying robots, such as the DelFly flapping wing MAV. In 2011-2012, he was a research fellow in the Advanced Concepts Team of the European Space Agency, where he studied topics such as optical flow based control algorithms for extraterrestrial landing scenarios. Currently, he is full professor at Delft University of Technology, the Netherlands, where he is the scientific lead of the Micro Air Vehicle Laboratory.
\end{IEEEbiography}

\clearpage
\pagestyle{empty}
\appendices

\section{Model details}\label{ap:moddet}

Table \ref{tab:flownetdet} outlines the full details of our version of FlowNetS. The name of the \textquote{conv}, \textquote{flow} and \textquote{predict\_flow} layer is abbreviated to \texttt{c}, \texttt{f} and \texttt{pf}, respectively. The change in the \texttt{pf} size from $3 \times 3$ to $1 \times 1$ helps interpretability of the analysis, but does bring the total receptive field size in the \texttt{c6} layer to 383 pixels as opposed to the original size of 511 pixels. The output of the \textquote{predict\_flow} layer is called \textquote{flow}.

In Table \ref{tab:perfcomp} a performance comparison between the slightly modified version of FlowNetS studied in this article and the original version of \textit{Dosovitskiy et al.} [17] can be found on the FlyingChairs [17] and MPI sintel [56] datasets. As shown, our version has a slightly worse, but comparable performance to the original version.

\begin{table*}[!htbp]
	\centering
	\caption{Full details of our version of FlowNetS.}
	\label{tab:flownetdet}
	\begin{tabular}{lccccccc}
		\toprule
		\textbf{Name} & \textbf{Kernel} & \textbf{Stride} & \textbf{Padding} & \textbf{Ch I/O} & \textbf{In Res} & \textbf{Out Res} & \textbf{Input}         \\
		\midrule
		conv1         & 7x7             & 2               & 3                & 6/64            & 512x384         & 256x192          & Images                 \\
		conv2         & 5x5             & 2               & 2                & 64/128          & 256x192         & 128x96           & conv1                  \\
		conv3\_0         & 5x5             & 2               & 2                & 128/256         & 128x96          & 64x48            & conv2                  \\
		conv3\_1      & 3x3             & 1               & 1                & 256/256         & 64x48           & 64x48            & conv3\_0                  \\
		conv4\_0         & 3x3             & 2               & 1                & 256/512         & 64x48           & 32x24            & conv3\_1               \\
		conv4\_1      & 3x3             & 1               & 1                & 512/512         & 32x24           & 32x24            & conv4\_0                  \\
		conv5\_0         & 3x3             & 2               & 1                & 512/512         & 32x24           & 16x12            & conv4\_1               \\
		conv5\_1      & 3x3             & 1               & 1                & 512/512         & 16x12           & 16x12            & conv5\_0                  \\
		conv6\_0         & 3x3             & 2               & 1                & 512/1024        & 16x12           & 8x6              & conv5\_1               \\
		conv6\_1      & 3x3             & 1               & 1                & 1024/1024       & 8x6             & 8x6              & conv6\_0                  \\
		\midrule
		predict\_flow6         & 1x1             & 1               & 1                & 1024/2          & 8x6             & 8x6              & conv6\_1               \\
		\midrule
		upconv5       & 4x4             & 2               & 1                & 1024/512        & 8x6             & 16x12            & conv6\_1               \\
		predict\_flow5         & 1x1             & 1               & 1                & 1026/2          & 16x12           & 16x12            & upconv5+conv5\_1+flow6 \\
		upconv4       & 4x4             & 2               & 1                & 1026/256        & 16x12           & 32x24            & upconv5+conv5\_1+flow6 \\
		predict\_flow4         & 1x1             & 1               & 1                & 770/2           & 32x24           & 32x24            & upconv4+conv4\_1+flow5 \\
		upconv3       & 4x4             & 2               & 1                & 770/128         & 32x24           & 64x48            & upconv4+conv4\_1+flow5 \\
		predict\_flow3         & 1x1             & 1               & 1                & 386/2           & 64x48           & 64x48            & upconv3+conv3\_1+flow4 \\
		upconv2       & 4x4             & 2               & 1                & 386/64          & 64x48           & 128x96           & upconv3+conv3\_1+flow4 \\
		predict\_flow2         & 1x1             & 1               & 1                & 192/2           & 128x96          & 128x96           & upconv2+conv2+flow3   \\
		\bottomrule
	\end{tabular}%
\end{table*}

\begin{table*}[!htbp]
	\centering
	\caption{Performance comparison between the original version of FlowNetS and ours on the MPI-Sintel [56] and FlyingChairs [17] datasets.}
	\label{tab:perfcomp}
	\resizebox{\textwidth}{!}{%
		\begin{tabular}{lp{60mm}lll}
			\toprule
			\textbf{Model name} & \textbf{Model details} & \textbf{FlyingChairs test {[}EPE{]}} & \textbf{MPI Sintel clean train {[}EPE{]}} & \textbf{MPI Sintel Final train {[}EPE{]}} \\
			\midrule
			FlowNetS [17]           & Original               & 2,71                             & 4,50                                      & 5,45                                      \\
			FlowNetS-ours           & ReLu activation function, \texttt{pf} layers with 1x1 kernels and no bias term, 300K training iterations, no data augmentation between frames                 & 3,10                             & 5,06                                      & 5,81                                     \\
			\bottomrule
		\end{tabular}%
	}
\end{table*}

\section{Grid search parameters}
\label{ap:params}

The parameter ranges used for the translation gridsearch are shown in Table \ref{tab:gridtrans}, while Table \ref{tab:fittrans} contains the parameters used for the spectral Gabor response profile fitting. 

The parameter ranges used for the dilation gridsearch can found in Table \ref{tab:griddil}. Note that due to rotational symmetry, the initial orientation $\theta$ only varies from 0 to 170 degrees. 

The parameters used for the rotation gridsearch are shown in Table \ref{tab:gridrot}. For this gridsearch, $\theta$ is also constrained from 0 to 170 degrees due to rotational symmetry. Note that the half spatial wavelength $\lambda /2$ can be transformed to spatial frequency $F_0$ using the relation $F = 1/2\lambda$.

\begin{table}[!htbp]
	\centering
	\caption{Parameter ranges used for the translating plane wave gridsearch.}
	\label{tab:gridtrans}
	\resizebox{.9\linewidth}{!}{
		\begin{tabular}{lll}
			\toprule
			\textbf{Parameter}       & \textbf{Unit} &\textbf{Range {[}start, stop, step size{]}} \\
			\midrule
			$\lambda /2$     & pixels & {[}16, 800, 16{]}                             \\
			$\theta$                 & degrees &{[}0, 350, 10{]}                              \\
			$f_{t}$                    & cycles per frame &{[}0.0, 0.5, 0.01{]}                          \\
			$\varphi$ & degrees &{[}-180, 170, 10{]}   \\
			\bottomrule                       
	\end{tabular}}
\end{table}

\begin{table}[!htbp]
	\centering
	\caption{Parameter ranges used for the Gabor spectral profile fitting process.}
	\label{tab:fittrans}
	\resizebox{.9\linewidth}{!}{
		\begin{tabular}{lll}
			\toprule
			\textbf{Parameter}       & \textbf{Unit} &\textbf{Range {[}start, stop, number of points{]}} \\
			\midrule
			$\lambda /2$  & cycles per pixel &  {[}16, 800, 50{]}                             \\
			$\theta$                 & degrees &{[}0, 350, 36{]}                              \\
			$f_{t}$                    & cycles per frame &{[}-0.5, 0.5, 50{]}                          \\
			\bottomrule                       
	\end{tabular}}
\end{table}

\begin{table}[!htbp]
	\centering
	\caption{Parameter ranges used for the dilating wave gridsearch.}
	\label{tab:griddil}
	\resizebox{.9\linewidth}{!}{
		\begin{tabular}{lll}
			\toprule
			\textbf{Parameter}       & \textbf{Unit} &\textbf{Range {[}start, stop, step size{]}} \\
			\midrule
			$\lambda /2$     & pixels & {[}50, 400, 10{]}                             \\
			$\theta$                 & degrees &{[}0, 170, 10{]}                              \\
			$s_f$                    & - &{[}0.5, 2.0, 0.1{]}                          \\
			$\varphi$ & degrees &{[}-180, 170, 10{]}   \\
			\bottomrule                       
	\end{tabular}}
\end{table}

\begin{table}[!htbp]
	\centering
	\caption{Parameter ranges used for the rotation gridsearch. The angular velocity $\omega$ is limited between $-0.5$ and $0.5$ cycles per sample which corresponds to $-\frac{1}{2}\pi$ and $\frac{1}{2}\pi$ radians per frame respectively.}
	\label{tab:gridrot}
	\resizebox{.9\linewidth}{!}{
		\begin{tabular}{lll}
			\toprule
			\textbf{Parameter}       & \textbf{Unit} &\textbf{Range {[}start, stop, step size{]}} \\
			\midrule
			$\lambda /2$     & pixels & {[}50, 400, 10{]}                             \\
			$\theta$                 & degrees &{[}0, 170, 10{]}                              \\
			$\omega$                    & cycles per frame &{[}-0.5, 0.5, 0.1{]}                          \\
			$\varphi$ & degrees &{[}-180, 170, 10{]}   \\
			\bottomrule                       
	\end{tabular}}
\end{table}

\section{Generalizability to natural images}
\label{ap:natural}

To evaluate the generalizability of the fitted bank of translational Gabor filters to natural stimuli, we used the first 500 image pairs of the FlyingChairs dataset [17] and compared the response of the Gabor bank to that of the corresponding filters from the convolutional \texttt{c6} layer of FlowNetS [17].

In order to obtain the response of the Gabor filters to a pair of input images, we first rendered the filter bank so that they have the same receptive field as the convolutional filters in the \texttt{c6} layer (i.e., $383\times 383$). We convolved the rendered filters over the padded image pair so that the output has the same spatial dimensions as the input images; and then applied the corresponding gain, bias, and the ReLU non-linearity as in Eq. 5. Lastly, we passed the resulting activations through a series of average pooling operations (same strides and padding as in FlowNetS) to obtain feature maps of the same spatial resolution as those from \texttt{c6} filters. These maps are directly comparable and we used the mean absolute error (MAE) for this purpose. We normalized the MAE for each filter and image pair using the maximum activation, and compute the average MAE over all image pairs in the dataset under analysis with which either the Gabor or the \texttt{c6} filter present a non-zero response. 

\begin{figure}[t]
	\centering
	\includegraphics[width=1\linewidth]{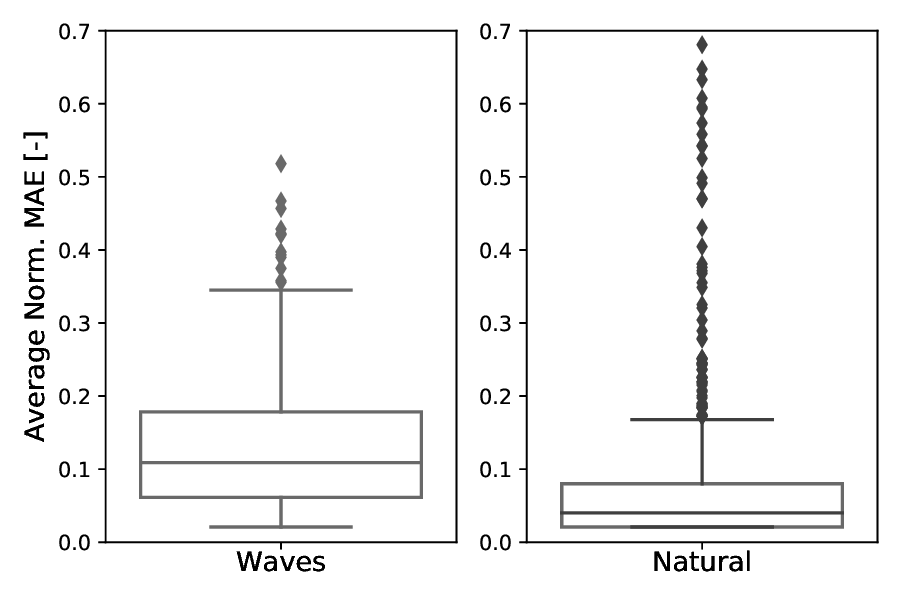}
	\caption{Average normalized MAE between the activations of convolutional filters in the \texttt{c6} layers and the corresponding fitted Gabor filters on datasets comprised of translational plane waves (left) and natural stimuli (right).}
	\label{fig:gen}
\end{figure}

Fig. \ref{fig:gen} shows the distributions of the errors on the natural stimuli from FlyingChairs and on a dataset comprised of the translating plane waves that maximally activate each of the fitted Gabor filters (i.e., 592 image pairs). These results show that, in both cases, the behavior of the fitted Gabor filters closely resembles that of the \texttt{c6} layer. The average error on the translating plane waves is 0.13 ($13\%$ of the maximal response), with the error distribution being characterized by only a few outliers (12) starting at $0.34$. On the natural stimuli, the average error is 0.08 ($8\%$ of the maximal response) with outliers (61) starting at 0.17. We believe the reason for the error being generally lower on the natural stimuli is due to the different image statistics between datasets, where the natural images come from the training data set.

Lastly, with this experiment, we also confirmed that the \texttt{c6} filters that we found silent during the fitting process (432 filters out of 1024) remain silent when presented with the natural stimuli from FlyingChairs. This, together with the error distributions in Fig. \ref{fig:gen}, validates the Gabor fitting process and the use of the translational plane waves as input images.

\end{document}